\documentclass[10pt,twocolumn,letterpaper]{article}

\usepackage{iccv}
\usepackage{times}
\usepackage{epsfig}
\usepackage{graphicx}
\usepackage{epstopdf}
\usepackage{amsmath}
\usepackage{amssymb}
\usepackage{algorithm}
\usepackage[noend]{algorithmic}
\usepackage{url}
\usepackage{dsfont}
\usepackage{epstopdf}
\usepackage{subfigure}
\usepackage{mathtools}
\usepackage{multirow}
\usepackage{footnote}

\usepackage[pagebackref=true,breaklinks=true,letterpaper=true,colorlinks,bookmarks=false]{hyperref}

\iccvfinalcopy 

\def\iccvPaperID{1060} 

\ificcvfinal\fi
\begin{document}
\title{RMPE: Regional Multi-Person Pose Estimation}

\author{Hao-Shu Fang$^{1}$\footnotemark[1],~~Shuqin Xie$^{1}$,~~Yu-Wing Tai$^{2}$,~~Cewu Lu$^{1}$\footnotemark[4]\\
$^{1}$Shanghai Jiao Tong University, China~~$^{2}$ Tencent YouTu\\
{\tt\small fhaoshu@gmail.com qweasdshu@sjtu.edu.cn yuwingtai@tencent.com lucewu@sjtu.edu.cn}}

\maketitle
\renewcommand{\thefootnote}{\fnsymbol{footnote}}
\footnotetext[1]{part of this work was done when Hao-Shu Fang was an student intern in Tencent}
\footnotetext[4]{corresponding author is Cewu Lu}
\footnotetext[2]{\href{https://cvsjtu.wordpress.com/rmpe-regional-multi-person-pose-estimation/}{https://cvsjtu.wordpress.com/rmpe-regional-multi-person-pose-estimation/}}
\begin{abstract}
Multi-person pose estimation in the wild is challenging. Although state-of-the-art human detectors have demonstrated good performance, small errors in localization and recognition are inevitable. These errors can cause failures for a single-person pose estimator (SPPE), especially for methods that solely depend on human detection results. In this paper, we propose a novel regional multi-person pose estimation (RMPE) framework to facilitate pose estimation in the presence of inaccurate human bounding boxes. Our framework consists of three components: Symmetric Spatial Transformer Network (SSTN), Parametric Pose Non-Maximum-Suppression (NMS), and Pose-Guided Proposals Generator (PGPG). Our method is able to handle inaccurate bounding boxes and redundant detections, allowing it to achieve ${\bf 76.7}$ mAP on the MPII (multi person) dataset\cite{andriluka14cvpr}. Our model and source codes are made \textbf{publicly available}.\footnotemark[2].
\end{abstract}


\section{Introduction}
\label{sec:Introduction}
Human pose estimation is a fundamental challenge for computer vision. In practice, recognizing the pose of multiple persons in the wild is a lot more challenging than recognizing the pose of a single person in an image~\cite{sapp2010cascaded,sun2012conditional,ladicky2013human,newell2016stacked,wei2016convolutional}. Recent attempts approach this problem by using either a two-step framework~\cite{pishchulin2012articulated,gkioxari2014using}  or
a part-based framework ~\cite{chen2015parsing,pishchulin16cvpr,insafutdinov16ariv}.
The two-step framework first detects human bounding boxes and then estimates the pose within each box independently. The part-based framework first detects body parts independently and then assembles the detected body parts to form multiple human poses. Both frameworks have their advantages and disadvantages. In the two-step framework, the accuracy of pose estimation highly depends on the quality of the detected bounding boxes. In the part-based framework, the assembled human poses are ambiguous when two or more persons are too close together. Also, part-based framework loses the capability to recognize body parts from a global pose view due to the mere utilization of second-order body parts dependence.

\begin{figure*}[hbt]
\centering
\begin{tabular}{@{\hspace{0mm}}c@{\hspace{1mm}}c@{\hspace{1mm}}c@{\hspace{1mm}}c@{\hspace{1mm}}c}
\includegraphics[width=0.47\linewidth]{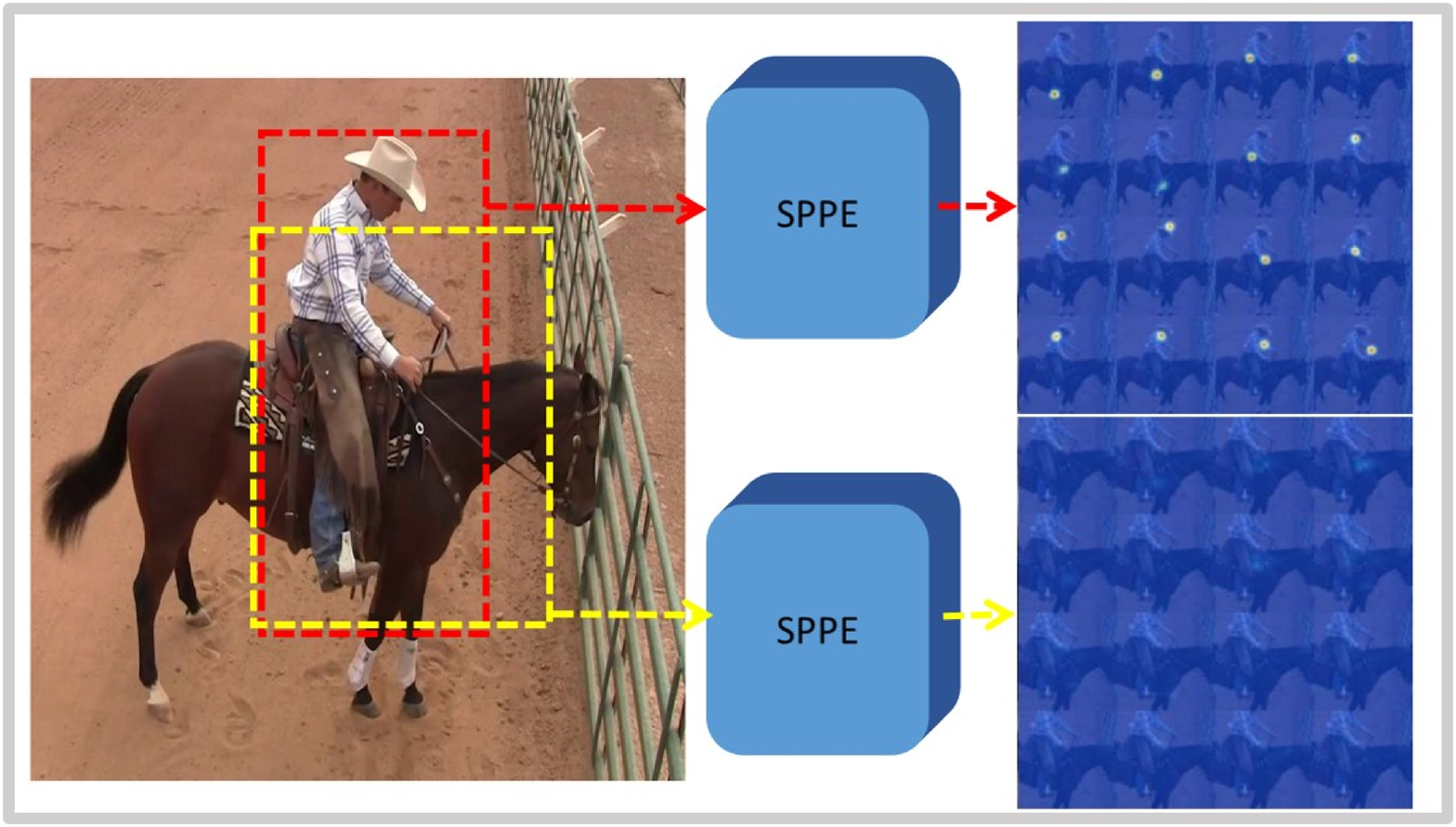}&
\includegraphics[width=0.47\linewidth]{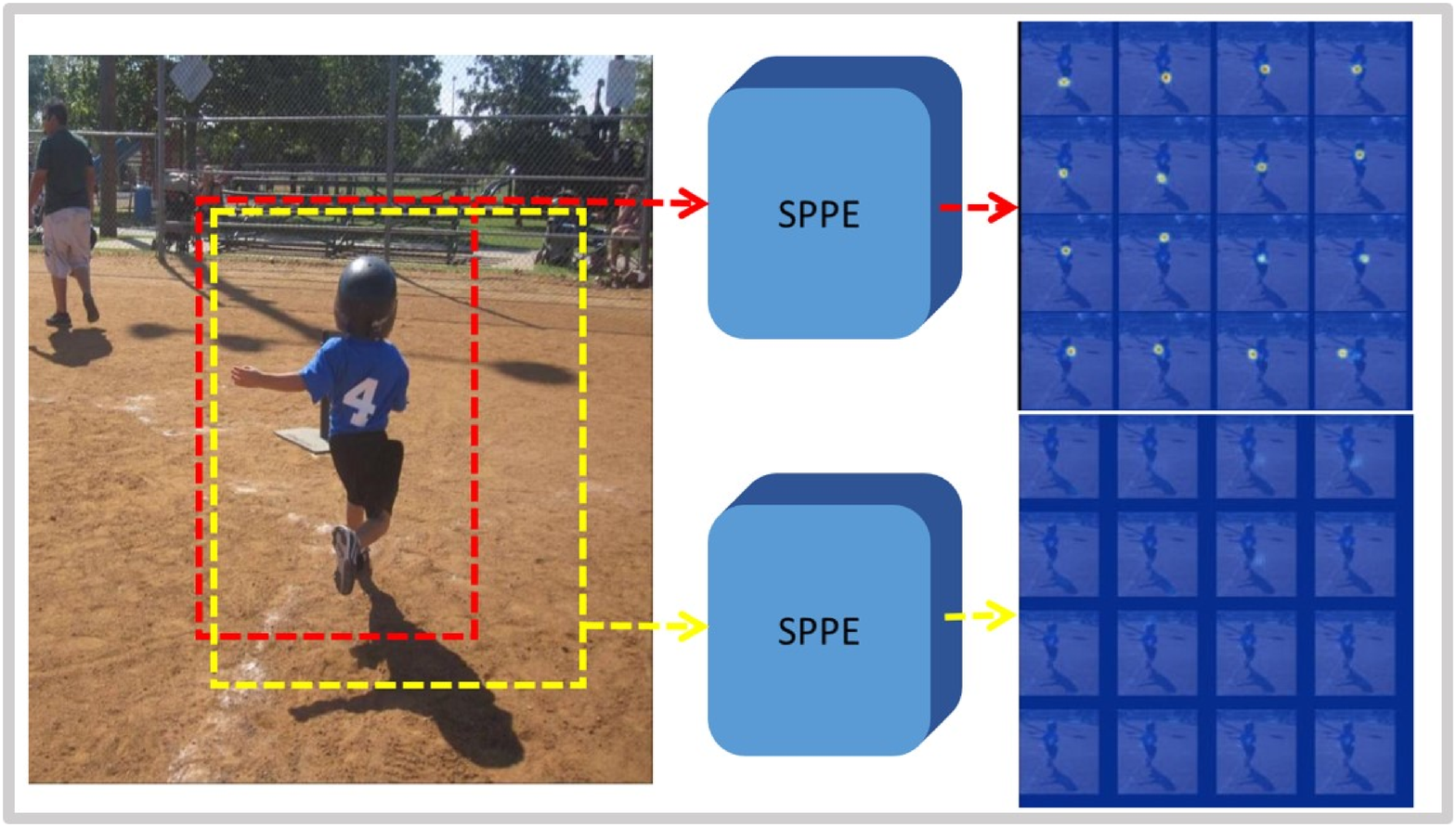}
\\
\end{tabular}
\caption{Problem of bounding box localization errors. The red boxes are the ground truth bounding boxes, and the yellow boxes are detected bounding boxes with $IoU > 0.5$. The heatmaps are the outputs of SPPE~\cite{newell2016stacked} corresponding to the two types of boxes. The corresponding body parts are not detected in the heatmaps of the yellow boxes. Note that with $IoU > 0.5$, the yellow boxes are considered as ``correct'' detections. However, human poses are not detected even with the ``correct'' bounding boxes.}
\vspace{-0.1in}
\label{fig:loc_error}
\end{figure*}

\begin{figure}[bt]
\centering
\begin{tabular}{@{\hspace{0mm}}c@{\hspace{1mm}}c@{\hspace{1mm}}c@{\hspace{1mm}}c@{\hspace{1mm}}c}
\includegraphics[width=0.9\linewidth]{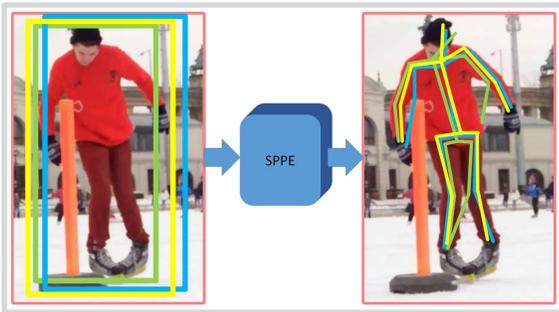}&
\end{tabular}
\caption{Problem of redundant human detections. The left image shows the detected bounding boxes; the right image shows the estimated human poses. Because each bounding box is operated on independently, multiple poses are detected for a single person.}
\vspace{-0.1in}
\label{fig:reco_error}
\end{figure}

Our approach follows the two-step framework. We aim to detect accurate human poses even when given inaccurate bounding boxes. To illustrate the problems of previous approaches, we applied the state-of-the-art object detector Faster-RCNN \cite{ren2015faster} and the SPPE Stacked Hourglass model \cite{newell2016stacked}. Figure \ref{fig:loc_error} and Figure \ref{fig:reco_error} show two major problems: the localization error problem and the redundant detection problem. In fact, SPPE is rather vulnerable to bounding box errors. Even for the cases when the bounding boxes are considered as correct with $IoU > 0.5$, the detected human poses can still be wrong. Since SPPE produces a pose for each given bounding box, redundant detections result in redundant poses.

To address the above problems, a regional multi-person pose estimation (RMPE) framework is proposed. Our framework improves the performance of SPPE-based human pose estimation algorithms. We have designed a new symmetric spatial transformer network (SSTN) which is attached to the SPPE to extract a high-quality single person region from an inaccurate bounding box. A novel parallel SPPE branch is introduced to optimize this network. To address the problem of redundant detection, a parametric pose NMS is introduced. Our parametric pose NMS eliminates redundant poses by using a novel pose distance metric to compare pose similarity. A data-driven approach is applied to optimize the pose distance parameters. Lastly, we propose a novel pose-guided human proposal generator (PGPG) to augment training samples. By learning the output distribution of a human detector for different poses, we can simulate the generation of human bounding boxes, producing a large sample of training data.

Our RMPE framework is general and is applicable to different human detectors and single person pose estimators. We applied our framework on the MPII (multi-person) dataset~\cite{andriluka14cvpr}, where it outperforms the state-of-the-art methods and achieves $76.7$ mAP. We have also conducted ablation studies to validate the effectiveness of each proposed component of our framework. Our model and source codes are made publicly available to support reproducible research.

\section{Related Work}\label{sec:related_work}

\subsection{Single Person Pose Estimation}
In single person pose estimation, the pose estimation problem is simplified by only attempting to estimate the pose of a single person, and the person is assumed to dominate the image content. Conventional methods considered pictorial structure models. For example, tree models~\cite{ wang2008multiple,sapp2010cascaded,zhang2009efficient,wang2013beyond} and random forest models ~\cite{sun2012conditional,dantone2013human} have demonstrated to be very efficient in human pose estimation.
Graph based models such as random field models~\cite{kiefel2014human} and dependency graph models~\cite{hara2013computationally} have also been widely investigated in the literature
~\cite{gupta2008context,sun2012efficient,ladicky2013human,pishchulin2013strong}.

More recently, deep learning has become a promising technique in object/face recognition, and human pose estimation is of no exception. Representative works include DeepPose (Toshev \emph{et al})~\cite{toshev2014deeppose}, DNN based models~\cite{Ouyang_2014_CVPR,fan2015combining} and various CNN based models~\cite{jain2013learning,tompson2014joint,newell2016stacked,belagiannis2016recurrent,wei2016convolutional}.
Apart from simply estimating a human pose, some studies ~\cite{dong2014towards,park2015attributed} consider human parsing
and pose estimation simultaneously. For single person pose estimation, these methods could perform well only when the person has been correctly located. However, this assumption is not always satisfied.

\begin{figure*}[hbt]
\centering
\begin{tabular}{@{\hspace{0mm}}c@{\hspace{1mm}}c@{\hspace{1mm}}c@{\hspace{1mm}}c@{\hspace{1mm}}c}
\includegraphics[width=0.95\linewidth]{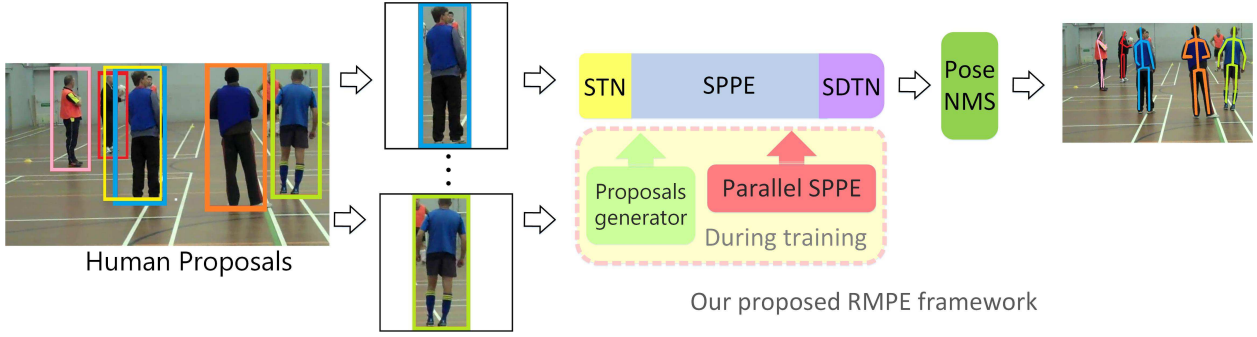}
\end{tabular}
\caption{Pipeline of our RMPE framework. Our \textbf{Symmetric STN} consists of \textbf{STN} and \textbf{SDTN} which are attached before and after the SPPE. The \textbf{STN} receives human proposals and the \textbf{SDTN} generates pose proposals. The \textbf{Parallel SPPE} acts as an extra regularizer during the training phase. Finally, the \textbf{parametric Pose NMS (p-Pose NMS)} is carried out to eliminate redundant pose estimations. Unlike traditional training, we train the SSTN+SPPE module with images generated by \textbf{PGPG}.}
\vspace{-0.1in}
\label{fig:framework}
\end{figure*}

\subsection{Multi Person Pose Estimation}
\noindent{\bf Part-based Framework}
Representative works on part-based framework \cite{chen2015parsing,gkioxari2014using,Iqbal_ECCVw2016,pishchulin16cvpr,insafutdinov16ariv} are reviewed. Chen \emph{et al}.  presented an approach to parse largely occluded people by graphical model which models humans as flexible compositions of body parts \cite{chen2015parsing}. Gkiox \emph{et al} used k-poselets to jointly detect people and predict locations of human poses \cite{gkioxari2014using}. The final pose localization is predicted by a weighted average of all activated poselets. Pishchulin \emph{et al}. proposed DeepCut to first detect all body parts, and then label and assemble these parts via integral linear programming\cite{pishchulin16cvpr}. A stronger part detector based on ResNet\cite{he2016deep} and a better incremental optimization strategy is proposed by Insafutdinov \emph{et al} \cite{insafutdinov16ariv}. While part-based methods have demonstrated good performance, their body-part detectors can be vulnerable since only small local regions are considered.

\vspace{2mm}
\noindent{\bf Two-step Framework}
Our work follows the two-step framework \cite{pishchulin2012articulated, gkioxari2014using}. In our work, we use a CNN based SPPE method to estimate poses, while Pishchulin \emph{et al}. \cite{pishchulin2012articulated} used conventional pictorial structure models for pose estimation. In particular, Insafutdinov \emph{et al} \cite{insafutdinov16ariv} propose a similar two-step pipeline which uses the Faster R-CNN as their human detector and a unary DeeperCut as their pose estimator. Their method can only achieve  $51.0$ in mAP on MPII dataset, while ours can achieve $76.7$ mAP. With the development of object detection and single person pose estimation, the two-step framework can achieve further advances in its performance. Our paper aims to solve the problem of imperfect human detection in the two-step framework in order to maximize the power of SPPE.


\section{Regional Multi-person Pose Estimation}


The pipeline of our proposed RMPE is illustrated in Figure \ref{fig:framework}. The human bounding boxes obtained by the human detector are fed into the ``\textbf{Symmetric STN} + SPPE'' module, and the pose proposals are generated automatically. The generated pose proposals are refined by \textbf{parametric Pose NMS} to obtain the estimated human poses. During the training, we introduce ``\textbf{Parallel SPPE}'' in order to avoid local minimums and further leverage the power of SSTN. To augment the existing training samples, a \textbf{pose-guided proposals generator (PGPG)} is designed. In the following sections, we present the three major components of our framework.

\begin{figure*}[hbt]
\centering
\begin{tabular}{@{\hspace{0mm}}c@{\hspace{1mm}}c@{\hspace{1mm}}c@{\hspace{1mm}}c@{\hspace{1mm}}c}
\includegraphics[width=0.95\linewidth]{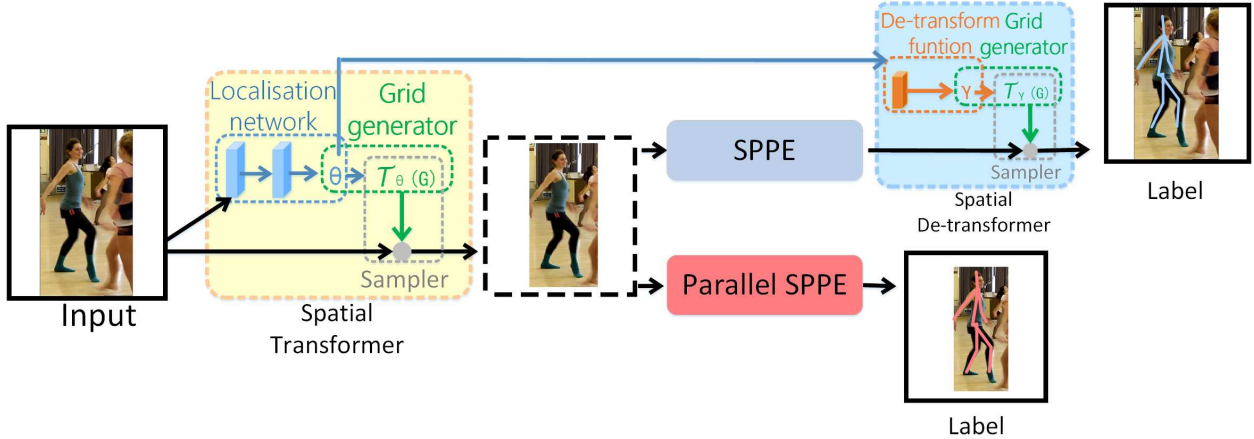}
\end{tabular}
\caption{An illustration of our symmetric STN architecture and our training strategy with parallel SPPE. The STN used was developed by Jaderberg \emph{et al.}~\cite{jaderberg2015spatial}. Our SDTN takes a parameter $\boldsymbol{\theta}$, generated by the localization net and computes the $\boldsymbol{\gamma}$ for de-transformation. We follow the grid generator and sampler~\cite{jaderberg2015spatial} to extract a human-dominant region. For our parallel SPPE branch, a center-located pose label is specified. We freeze the weights of all layers of the parallel SPPE to encourage the STN to extract a dominant single person proposal.
}
\vspace{-0.1in}
\label{fig:sstn+para}
\end{figure*}

\subsection{Symmetric STN and Parallel SPPE}
Human proposals provided by human detectors are not well-suited to SPPE. This is because SPPE is specifically trained on single person images and is very sensitive to localisation errors. It has been shown that small translation or cropping of human proposals can significantly affect performance of SPPE \cite{newell2016stacked}. Our symmetric STN $+$ parallel SPPE was introduced to enhance SPPE when given imperfect human proposals. The module of our SSTN and parallel SPPE is shown in Figure \ref{fig:sstn+para}.

\vspace{2mm}
\noindent{\bf STN and SDTN} The spatial transformer network \cite{jaderberg2015spatial}(STN) has demonstrated excellent performance
in selecting region of interests automatically. In this paper, we use the STN to extract high quality dominant human proposals. Mathematically, the STN performs a 2D affine transformation which can be expressed as
\begin{equation}\label{eq1}
\left(
\begin{matrix}
   x_{i}^{s} \\
   y_{i}^{s}
  \end{matrix}
  \right) =
\left[
  \begin{matrix}
  \boldsymbol{\theta_{1}} & \boldsymbol{\theta_{2}} & \boldsymbol{\theta_{3}}
  \end{matrix}
\right]
\left(
\begin{matrix}
   x_{i}^{t} \\
   y_{i}^{t}  \\
   1
  \end{matrix}
\right),
\end{equation}
where $\boldsymbol{\theta_{1}}$, $\boldsymbol{\theta_{2}}$ and $\boldsymbol{\theta_{3}}$ are vectors in $\mathbb{R}^2$. $\{x_{i}^{s},y_{i}^{s}\}$ and $\{x_{i}^{t},y_{i}^{t}\}$ are the coordinates before and after transformation, respectively. After SPPE, the resulting pose is mapped into the original human proposal image. Naturally, a spatial de-transformer network (SDTN) is required to remap the estimated human pose back to the original image coordinate. The SDTN computes the $\boldsymbol{\gamma}$ for de-transformation and generates grids based on $\boldsymbol{\gamma}$:
{\small
\begin{equation}\label{eq2}
\left(
\begin{matrix}
   x_{i}^{t} \\
   y_{i}^{t}
  \end{matrix}
  \right) =
\left[
  \begin{matrix}
  \boldsymbol{\gamma_{1}} & \boldsymbol{\gamma_{2}} & \boldsymbol{\gamma_{3}}
  \end{matrix}
\right]
\left(
\begin{matrix}
   x_{i}^{s} \\
   y_{i}^{s}  \\
   1
  \end{matrix}
\right)
\end{equation}
}
Since SDTN is an inverse procedure of STN, we can obtain the following:
{\small
\begin{equation}\label{eq5}
\left[
  \begin{matrix}
  \boldsymbol{\gamma_{1}} & \boldsymbol{\gamma_{2}}
  \end{matrix}
\right]
=
\left[
  \begin{matrix}
  \boldsymbol{\theta_{1}} & \boldsymbol{\theta_{2}}
  \end{matrix}
\right]^{-1}
\end{equation}
\begin{equation}\label{eq6}
\boldsymbol{\gamma_{3}}
=
-1\times\left[
  \begin{matrix}
  \boldsymbol{\gamma_{1}} & \boldsymbol{\gamma_{2}}
  \end{matrix}
\right]
\boldsymbol{\theta_{3}}
\end{equation}
}
To back propagate through SDTN, $\frac{\partial J(W,b)}{\partial\theta}$ can be derived as
{\small
\begin{equation}\label{eq7}
\begin{split}
\frac{\partial J(W,b)}{\partial\left[
  \begin{matrix}
  \boldsymbol{\theta_{1}} & \boldsymbol{\theta_{2}}
  \end{matrix}
\right]}
=
\frac{\partial J(W,b)}{\partial\left[
  \begin{matrix}
  \boldsymbol{\gamma_{1}} & \boldsymbol{\gamma_{2}}
  \end{matrix}
\right]}
\times
\frac{\partial \left[
  \begin{matrix}
  \boldsymbol{\gamma_{1}} & \boldsymbol{\gamma_{2}}
  \end{matrix}
\right]}{\partial\left[
  \begin{matrix}
  \boldsymbol{\theta_{1}} & \boldsymbol{\theta_{2}}
  \end{matrix}
\right]}\\
+
\frac{\partial J(W,b)}{\partial\boldsymbol{\gamma_{3}}}
\times
\frac{\partial \boldsymbol{\gamma_{3}}}{\partial\left[
  \begin{matrix}
  \boldsymbol{\gamma_{1}} & \boldsymbol{\gamma_{2}}
  \end{matrix}
\right]}
\times
\frac{\partial \left[
  \begin{matrix}
  \boldsymbol{\gamma_{1}} & \boldsymbol{\gamma_{2}}
  \end{matrix}
\right]}{\partial\left[
  \begin{matrix}
  \boldsymbol{\theta_{1}} & \boldsymbol{\theta_{2}}
  \end{matrix}
\right]}
\end{split}
\end{equation}
}
with respect to $\theta_{1}$ and $\theta_{2}$, and
{\small
\begin{equation}\label{eq8}
\frac{\partial J(W,b)}{\partial\boldsymbol{\theta_{3}}}
=
\frac{\partial J(W,b)}{\partial\boldsymbol{\gamma_{3}}}
\times
\frac{\partial\boldsymbol{\gamma_{3}}}{\partial\boldsymbol{\theta_{3}}}
\end{equation}
}
with respect to $\theta_{3}$.
$\frac{\partial \left[
  \begin{matrix}
  \boldsymbol{\gamma_{1}} & \boldsymbol{\gamma_{2}}
  \end{matrix}
\right]}{\partial\left[
  \begin{matrix}
  \boldsymbol{\theta_{1}} & \boldsymbol{\theta_{2}}
  \end{matrix}
\right]}$ and
$\frac{\partial\boldsymbol{\gamma_{3}}}{\partial\boldsymbol{\theta_{3}}}$ can be derived from Eqn. \eqref{eq5} and \eqref{eq6} respectively.

After extracting high quality dominant human proposal regions, we can utilize off-the-shelf SPPE for accurate pose estimation. In our training, the SSTN is fine-tuned together with our SPPE.

\paragraph{Parallel SPPE}
To further help STN extract good human-dominant regions, we add a parallel SPPE branch in the training phrase. This branch shares the same STN with the original SPPE, but the spatial de-transformer (SDTN) is omitted. The human pose label of this branch is specified to be centered. To be more specific, the output of this SPPE branch is directly compared to labels of center-located ground truth poses. We freeze all the layers of this parallel SPPE during the training phase. The weights of this branch are fixed and its purpose is to back-propagate center-located pose errors to the STN module. If the extracted pose of the STN is not center-located, the parallel branch will back-propagate large errors. In this way, we can help the STN focus on the correct area and extract high quality human-dominant regions. In the testing phase, the parallel SPPE is discarded. The effectiveness of our parallel SPPE will be verified in our experiments.

\vspace{2mm}
\noindent{\bf Discussions} The parallel SPPE can be regarded as a regularizer during the training phase. It helps to avoid a poor solution (local minimum) where the STN does not transform the pose to the center of extracted human regions. The likelihood of reaching a local minimum is increased because compensation from the SDTN will make the network generate fewer errors. These errors are necessary to train the STN. With the parallel SPPE, the STN is trained to move the human to the center of the extracted region to facilitate accurate pose estimation by SPPE.

It may seem intuitive to replace parallel SPPE with a center-located poses regression loss in the output of SPPE (before SDTN). However, this approach will degrade the performance of our system.  Although STN can partly transform the input, it is impossible to perfectly place the person at the same location as the label. The difference in coordinate space between the input and label of SPPE will largely impair its ability to learn pose estimation. This will cause the performance of our main branch SPPE to decrease. Thus, to ensure that both STN and SPPE can fully leverage their own power, a parallel SPPE with frozen weights is indispensable for our framework. The parallel SPPE always produces large errors for non-center poses to push the STN to produce a center-located pose, without affecting the performance of the main branch SPPE.

\subsection{Parametric Pose NMS}
Human detectors inevitably generate redundant detections, which in turn produce redundant pose estimations. Therefore, pose non-maximum suppression (NMS) is required to eliminate the redundancies. Previous methods \cite{burgos2013merging,chen2015parsing} are either not efficient or not accurate enough. In this paper, we propose a parametric pose NMS method. Similar to the previous subsection, the pose  $P_i$, with $m$ joints is denoted as $\{\langle k_i^{1} , c_i^{1} \rangle, \ldots,\langle k_i^{m} , c_i^{m} \rangle \}$, where $k_i^j$ and $c_i^j$ are the $j^{th}$ location and confidence score of joints respectively.

\vspace{2mm}
\noindent{\bf NMS scheme} We revisit pose NMS as follows: firstly, the most confident pose is selected as reference, and some poses close to it are subject to elimination by applying \emph{elimination criterion}. This process is repeated on the remaining poses set until redundant poses are eliminated and only unique poses are reported.

\vspace{2mm}
\noindent{\bf Elimination Criterion}  We need to define pose similarity in order to eliminate the poses which are too close and too similar to each others. We define a pose distance metric $d(P_{i}, P_{j}| \Lambda)$ to measure the pose similarity, and a threshold $\eta$ as elimination criterion, where $\Lambda$ is a parameter set of function $d(\cdot)$. Our elimination criterion can be written as follows:
\begin{equation}\label{eq:elimination_criterion}
 f(\ P_{i}, P_{j}|  \Lambda,  \eta) = \mathds{1}[d(P_{i}, P_{j}|  \Lambda, \lambda) \leq \eta]
\end{equation}
If $d(\cdot)$ is smaller than $\eta$, the output of $f(\cdot)$  should be $1$, which indicates that pose $P_{i}$ should be eliminated due to redundancy with reference pose $P_{j}$.

\paragraph{Pose Distance} Now, we present the distance function $d_{pose}(P_{i}, P_{j})$. We assume that the box for $P_{i}$ is $B_{i}$. Then we define a soft matching function
\begin{eqnarray}\label{eq:count_distance}
K_{Sim}( P_{i}, P_{j}| \sigma_1 ) = ~~~~~~~~~~~~~~~~~~~~~~~~~~~~~~~~~~~~~~~~~~~~~~ \nonumber\\
\begin{cases}
\sum_{n}  \tanh \frac{c_{i}^{n}}{\sigma_1} \cdot \tanh \frac{c_{j}^{n}}{\sigma_1}, & \text{if $k_{j}^{n}$ is within $\mathcal{B}(k_i^{n})$} \\
0 & \text{otherwise }
\end{cases}
\end{eqnarray}
where $\mathcal{B}(k_i^{n})$ is a box center at $k_i^{n}$, and each dimension of $\mathcal{B}(k_i^{n})$ is $1/10$ of the original box $B_{i}$. The $\tanh$ operation filters out poses with low-confidence scores. When two corresponding joints both have high confidence scores, the output will be close to 1. This distance softly counts the number of joints matching between poses.

The spatial distance between parts is also considered, which can be written as
\begin{equation}\label{eq:spatial_distance}
H_{Sim}( P_{i}, P_{j}| \sigma_2 ) = \sum_{n} \exp[- \frac{(k_{i}^{n}- k_{j}^{n})^2}{\sigma_2}]
\end{equation}

By combining Eqn \eqref{eq:count_distance} and \eqref{eq:spatial_distance}, the final distance function can be written as
\begin{equation}\label{eq:energy}
d(P_{i}, P_{j}|  \Lambda) = K_{Sim}( P_{i}, P_{j}| \sigma_1 ) +  \lambda H_{Sim}( P_{i}, P_{j}| \sigma_2 )
\end{equation}
where $\lambda$ is a weight balancing the two distances and $\Lambda = \{\sigma_1, \sigma_2, \lambda\}$. Note that the previous pose NMS \cite{chen2015parsing} set pose distance parameters and thresholds manually. In contrast, our parameters can be determined in a data-driven manner.

\vspace{2mm}
\noindent{\bf Optimization} Given the detected redundant poses, the four parameters in the eliminate criterion $f(\ P_{i}, P_{j}|\Lambda, \eta)$ are optimized to achieve the maximal mAP for the validation set. Since exhaustive search in a 4D space is intractable, we optimize two parameters at a time by fixing the other two parameters in an iterative manner. Once convergence is achieved, the parameters are fixed and will be used in the testing phase.

\subsection{Pose-guided Proposals Generator}
\paragraph{Data Augmentation} For the two-stage pose estimation, proper data augmentation is necessary to make the SSTN+SPPE module adapt to the 'imperfect' human proposals generated by the human detector. Otherwise, the module may not work properly in the testing phase for the human detector. An intuitive approach is to directly use bounding boxes generated by the human detector during the training phase. However, the human detector can only produce one bounding box for each person. By using the proposals generator, this quantity can be greatly increased. Since we already have the ground truth pose and an object detection bounding box for each person, we can generate a large sample of training proposals with the same distribution as the output of the human detector. With this technique, we are able to further boost the performance of our system.

\vspace{2mm}
\noindent{\bf Insight} We find that the distribution of the relative offset between the detected bounding box and the ground truth bounding box varies across different poses. To be more specific, there exists a distribution $P(\delta B | P)$, where $\delta B$ is the offset between the coordinates of a bounding box generated by human detector and the coordinates of the ground truth bounding box, and $P$ is the ground truth pose of a person. If we can model this distribution, we are able to generate many training samples that are  similar to human proposals generated by the human detector.

\begin{figure}[bt]
\begin{center}
   \includegraphics[width=1\linewidth]{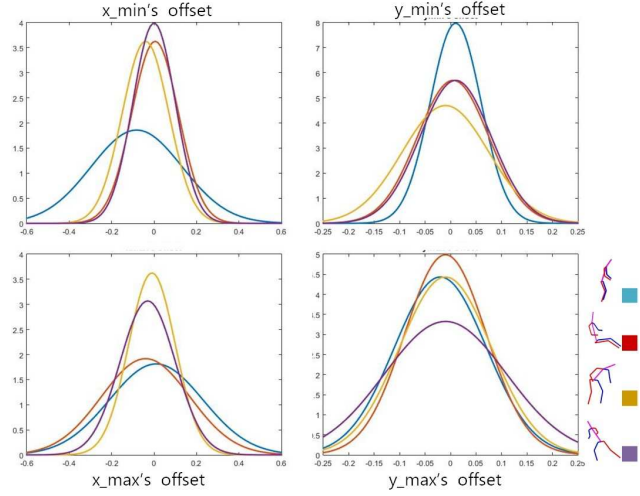}
\end{center}
   \caption{Gaussian distributions of bounding box offsets for several different atomic poses. More results are available in supplementary materials. Best viewed in color.}
\vspace{-0.1in}
\label{fig:pose-distribution}
\end{figure}

\vspace{2mm}
\noindent{\bf Implementation} To directly learn the distribution $P(\delta B | P)$ is difficult due to the variation of human poses. So instead, we attempt to learn the distribution $P(\delta B | atom(P))$, where $atom(P)$ denotes the atomic pose~\cite{yao2012recognizing} of $P$. We follow the method used by  Andriluka \emph{et al}~\cite{andriluka14cvpr} to learn the atomic poses. To derive the atomic poses from annotations of human poses, we first align all poses so that their torsos have the same length. Then we use the k-means algorithm to cluster our aligned poses, and the computed cluster centers form our atomic poses. Now for each person instance sharing the same atomic pose $a$, we calculate the offsets between its ground truth bounding box and detected bounding box. The offsets are then normalized by the corresponding side-length of ground truth bounding box in that direction. After these processes, the offsets form a frequency distribution, and we fit our data to a Gaussian mixture distribution. For different atomic poses, we have different Gaussian mixture parameters. We visualize some of the distributions and their corresponding clustered human poses in Figure~\ref{fig:pose-distribution}.

\vspace{2mm}
\noindent{\bf Proposals Generation} During the training phase of the SSTN+SPPE, for each annotated pose in the training sample we first look up the corresponding atomic pose $a$. Then we generate additional offsets by dense sampling according to $P(\delta B | a)$ to produce augmented training proposals.

\section{Experiments}
\label{Experiments}
The proposed method is qualitatively and quantitatively evaluated on two standard multi-person datasets with large occlusion cases: MPII \cite{andriluka14cvpr} and MSCOCO 2016 Keypoints Challenge dataset\cite{coco16}.


\subsection{Evaluation datasets}
\vspace{2mm}
\noindent{\bf MPII Multi-Person Dataset} The challenging benchmark MPII Human Pose (multi-person)\cite{andriluka14cvpr} consists of 3,844 training and 1,758 testing groups with both occluded and overlapped people. Moreover, it contains more than 28,000 training samples for single person pose estimation. We use all the training data in the single person dataset and 90\% of the multi-person training set to fine-tune the SPPE, leaving 10\% for validation.

\vspace{2mm}
\noindent{\bf MSCOCO Keypoints Challenge} We also evaluate our method on the MSCOCO Keypoints Challenge dataset\cite{coco16}. This dataset requires localization of person keypoints in challenging, uncontrolled conditions. It consists of 105,698 training and around 80,000 testing human instances. The training set contains over 1 million total labeled keypoints. The testing set are divided into four roughly equally sized splits: test-challenge, test-dev, test-standard, and test-reserve.

\subsection{Implementation details in testing}
In this paper, we use the VGG-based SSD-512 \cite{liu2015ssd} as our human detector, as it performs object detection effectively and efficiently. In order to guarantee that the entire person region will be extracted, detected human proposals are extended by $30\%$ along both the height and width directions. We use the stacked hourglass model~\cite{newell2016stacked} as the single person pose estimator because of its superior performance. For the STN network, we adopt the ResNet-18~\cite{he2016deep} as our localization network. Considering the memory efficiency, we use a smaller 4-stack hourglass network as the parallel SPPE.

\vspace{1mm}
To show that our framework is general and is applicable to different human detectors and pose estimators, we also do experiments by replacing the human detector with ResNet152 based Faster-RCNN~\cite{chen17implementation} and replacing the pose estimator with PyraNet~\cite{yang2017learning}. In this case, we adopt multi-scale testing for the human detection and use an input size of 320x256 for the PyraNet.

\begin{table*}[tbh]
\begin{center}
\begin{tabular}{c||c c c c c c c c}
\hline
 &Head & Shoulder & Elbow & Wrist & Hip & Knee  & Ankle & Total\\
 \hline
 \multicolumn{9}{ c }{full testing set} \\
Iqbal\&Gall, ECCVw16 \cite{Iqbal_ECCVw2016}& 58.4  & 53.9  & 44.5  & 35.0  & 42.2  & 36.7 & 31.1 & 43.1\\
DeeperCut, ECCV16 \cite{insafutdinov16ariv}& 78.4  & 72.5  & 60.2  & 51.0  & 57.2  & 52.0 & 45.4 & 59.5\\
Levinkov \emph{et al.}, CVPR17\cite{levinkov2017cvpr}& 89.8 & 85.2 & 71.8 & 59.6 & 71.1 & 63.0 & 53.5 & 70.6\\
Insafutdinov \emph{et al.}, CVPR17\cite{insafutdinov2017cvpr}& 88.8 & 87.0 & 75.9 & 64.9 & 74.2 & 68.8 & 60.5 & 74.3\\
Cao \emph{et al.}, CVPR17\cite{cao2017realtime} & 91.2 & 87.6 & 77.7 & 66.8 & 75.4 & 68.9 & 61.7 & 75.6\\
Newell \& Deng, NIPS17\cite{newell2017associative}& \textbf{92.1} & 89.3 & 78.9 & 69.8 & 76.2 & 71.6 & 64.7 & 77.5\\
\hline
\textbf{ours}& 88.4 & 86.5  & {78.6}  & 70.4  & 74.4  & {73.0} & {65.8} & {76.7}\\
\textbf{ours++}& 91.3 & \textbf{90.5}  & \textbf{84.0}  & \textbf{76.4}  & \textbf{80.3}  & \textbf{79.9} & \textbf{72.4} & \textbf{82.1}\\
\hline
\end{tabular} \vspace{0.10in}
\caption{Results on the MPII multi-person test set (mAP). ``++'' denotes using faster-rcnn with softnms~\cite{bodla2017soft} as human detector, PyraNet~\cite{yang2017learning} with input size 320x256 as pose estimator.} \label{tab:compare}
\end{center}
\end{table*}

\subsection{Results}

\begin{figure*}[hbt]
\centering
\begin{tabular}{@{\hspace{0mm}}c@{\hspace{1mm}}c@{\hspace{1mm}}c@{\hspace{1mm}}c}
\includegraphics[width=0.25\linewidth]{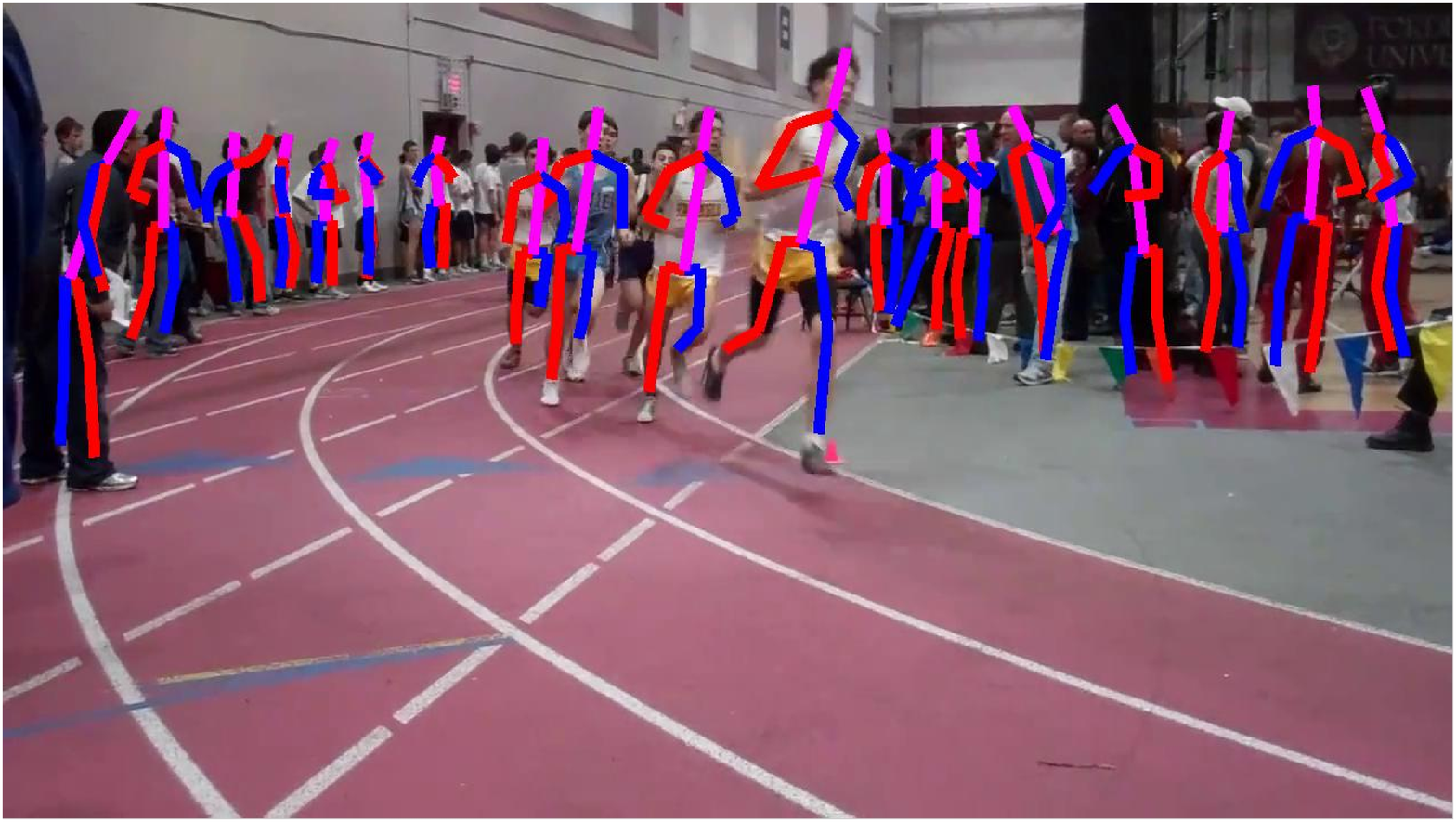}&
\includegraphics[width=0.25\linewidth]{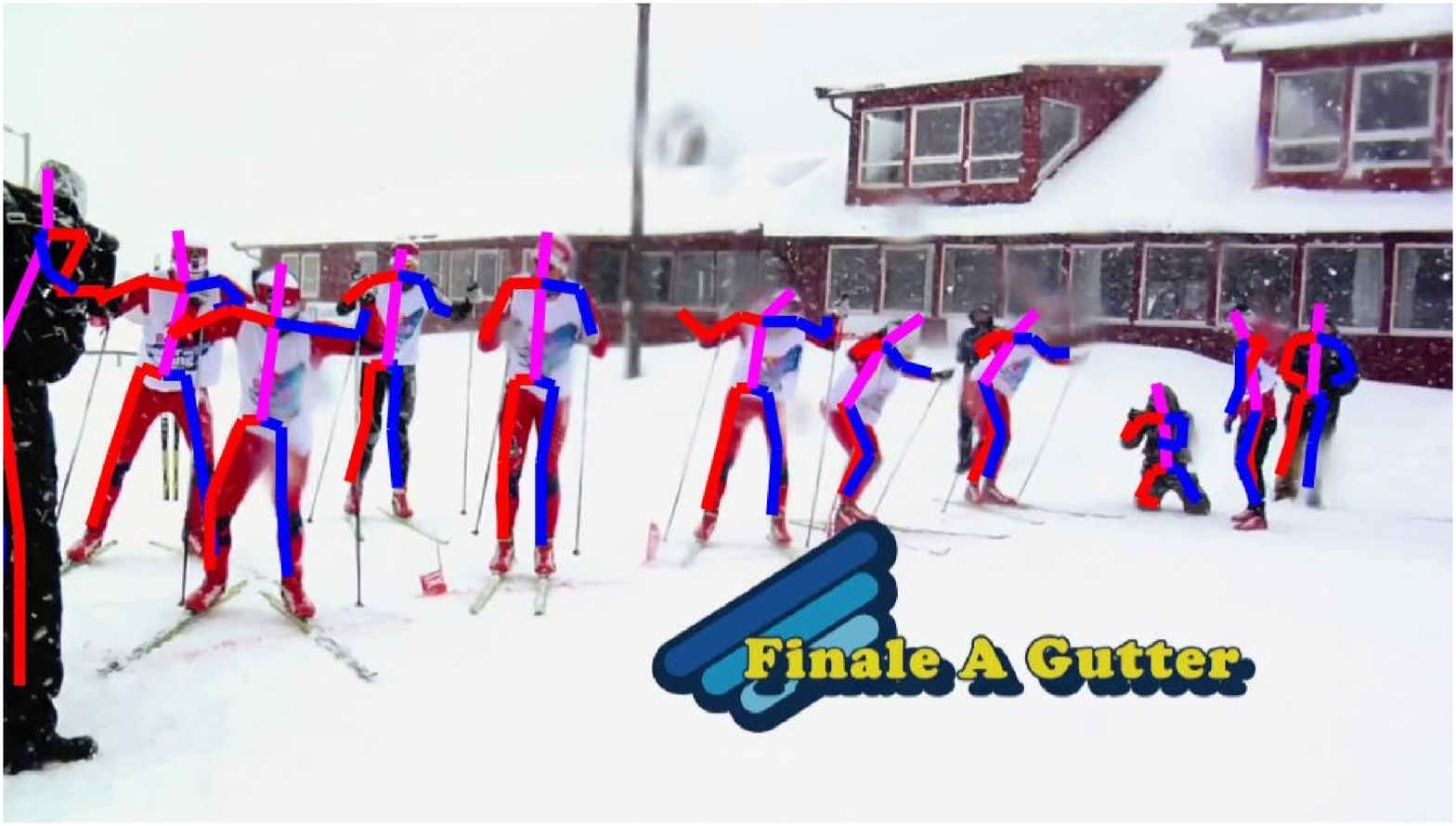}&
\includegraphics[width=0.25\linewidth]{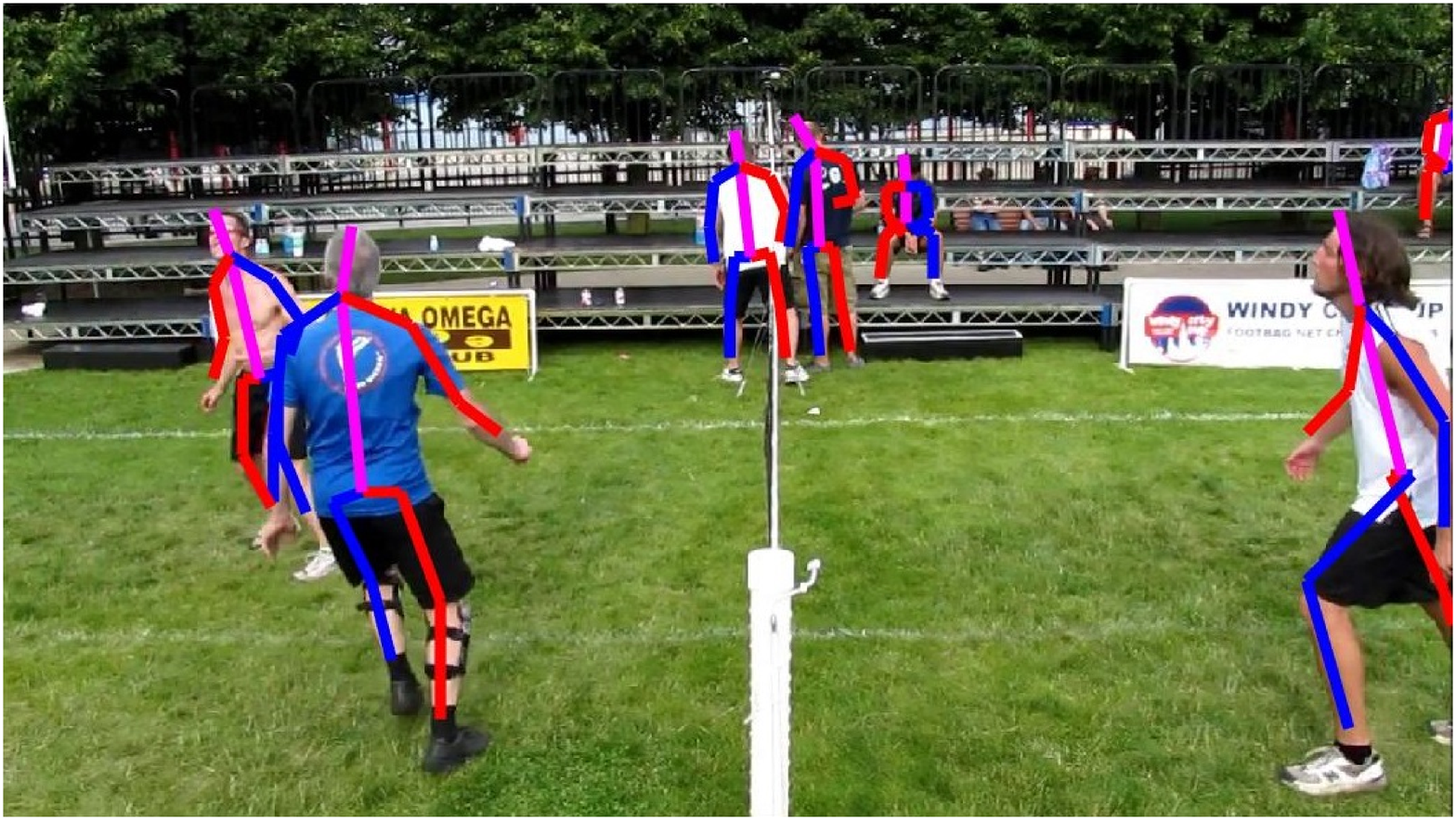}&
\includegraphics[width=0.25\linewidth]{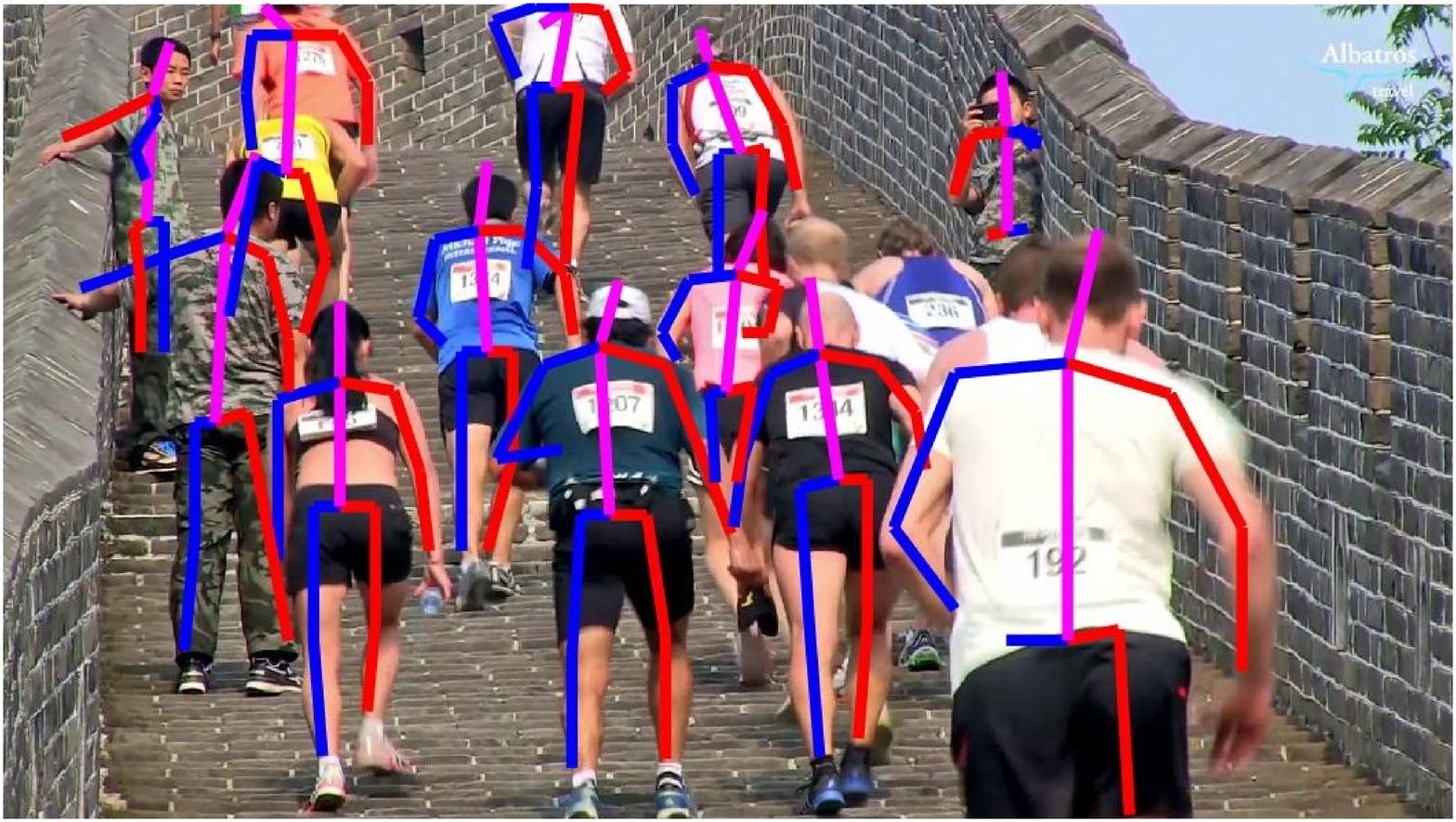}\\
\includegraphics[width=0.25\linewidth]{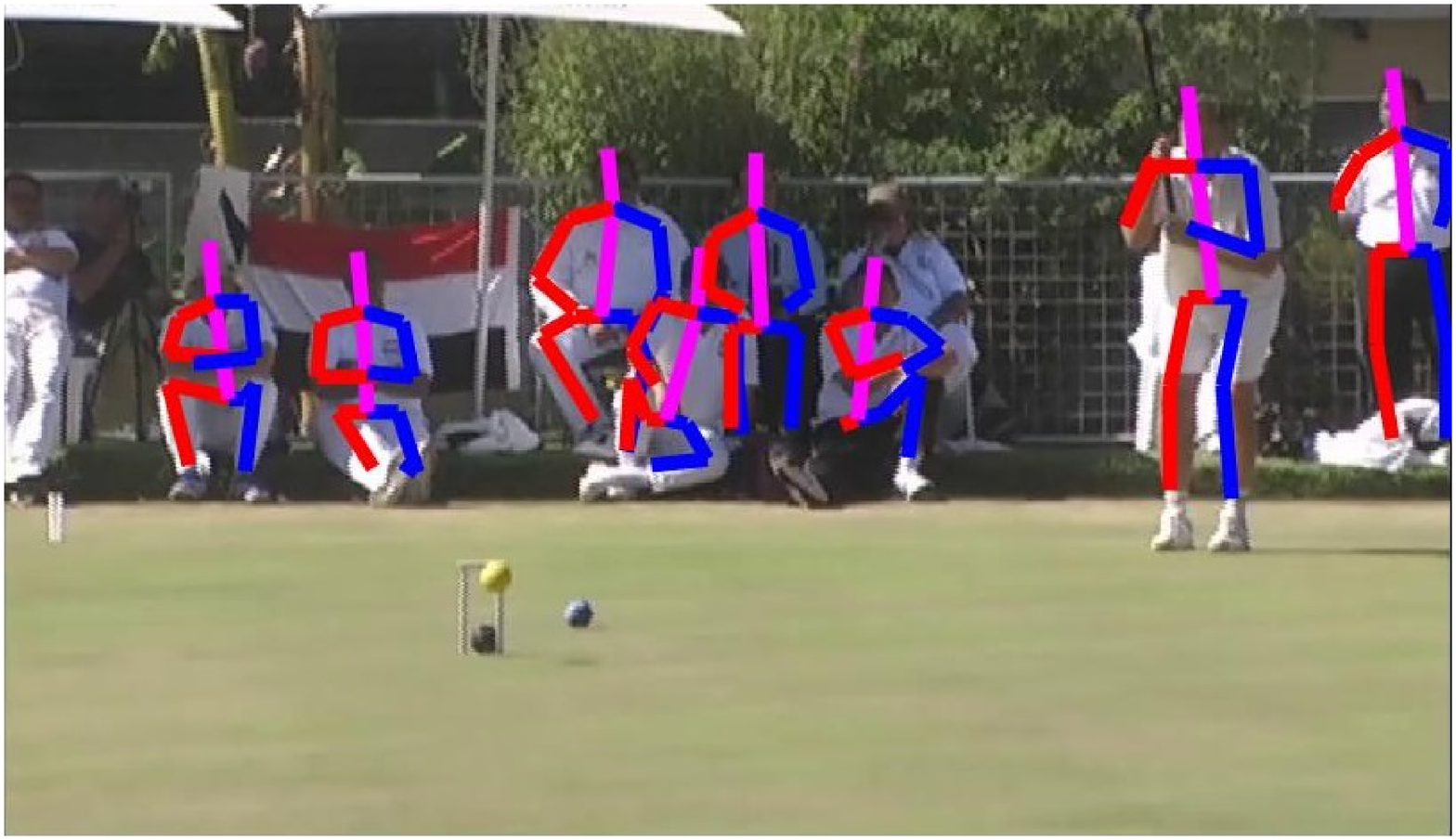}&
\includegraphics[width=0.25\linewidth]{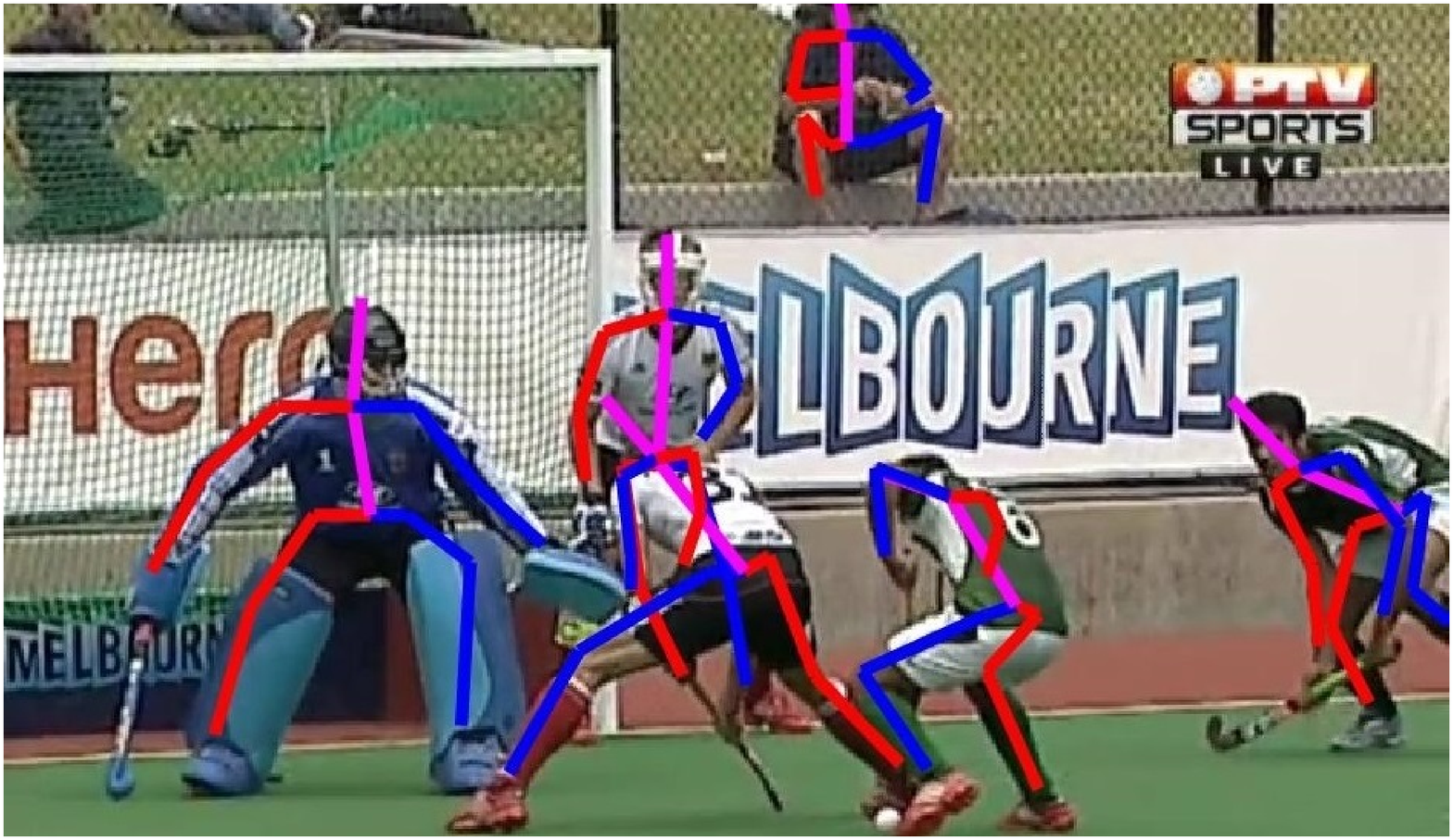}&
\includegraphics[width=0.25\linewidth]{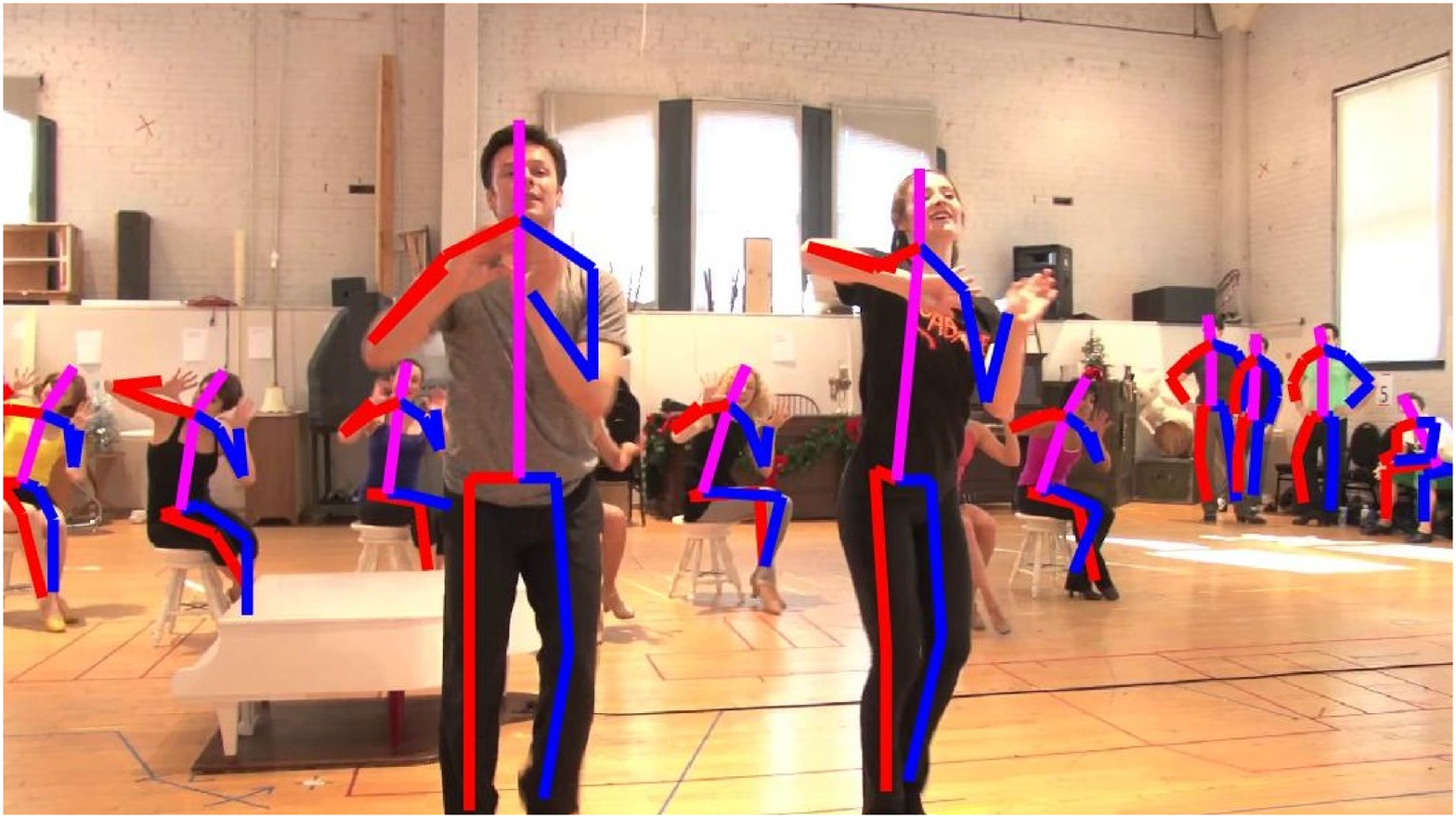}&
\includegraphics[width=0.25\linewidth]{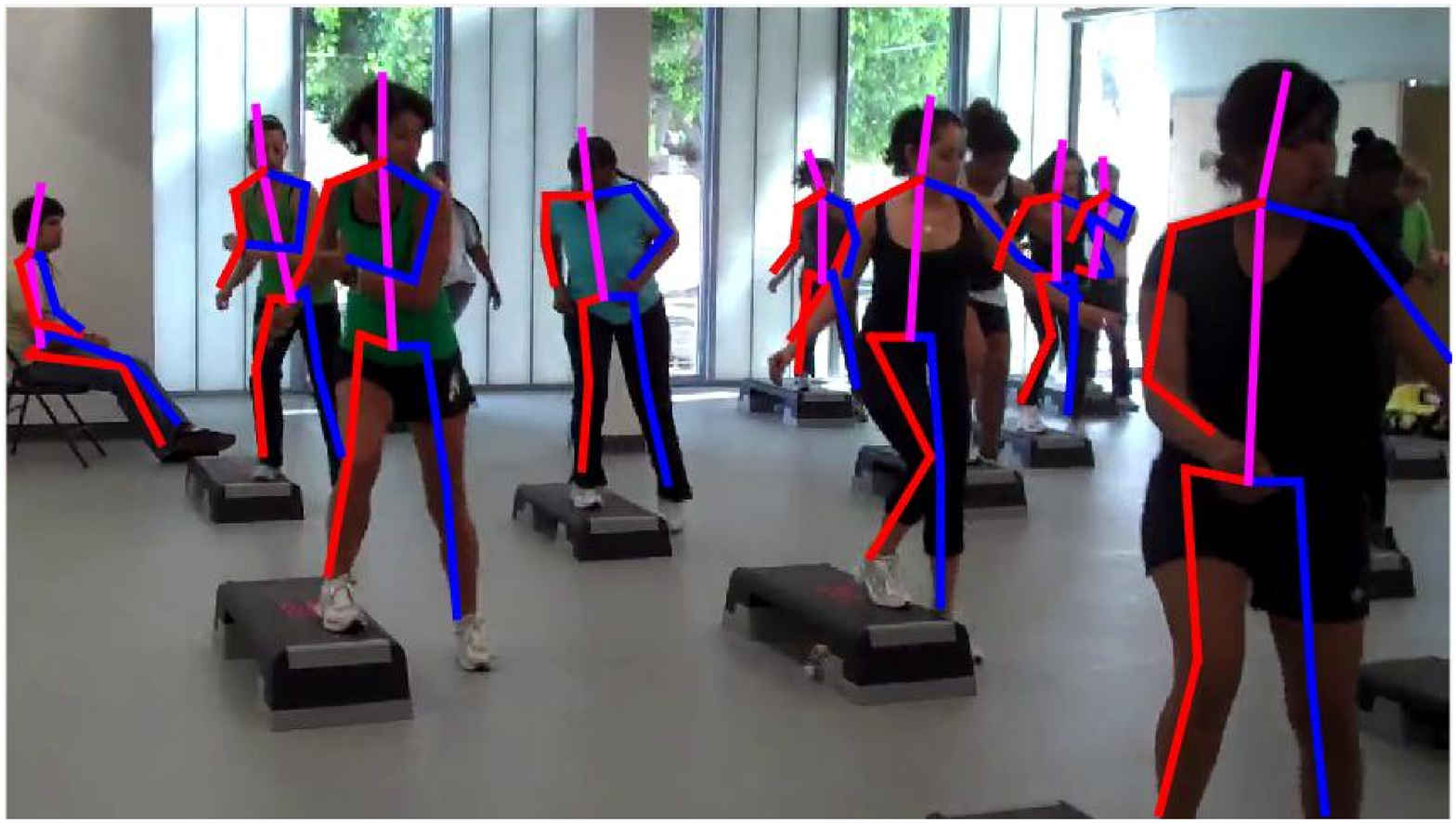}\\
\includegraphics[width=0.25\linewidth]{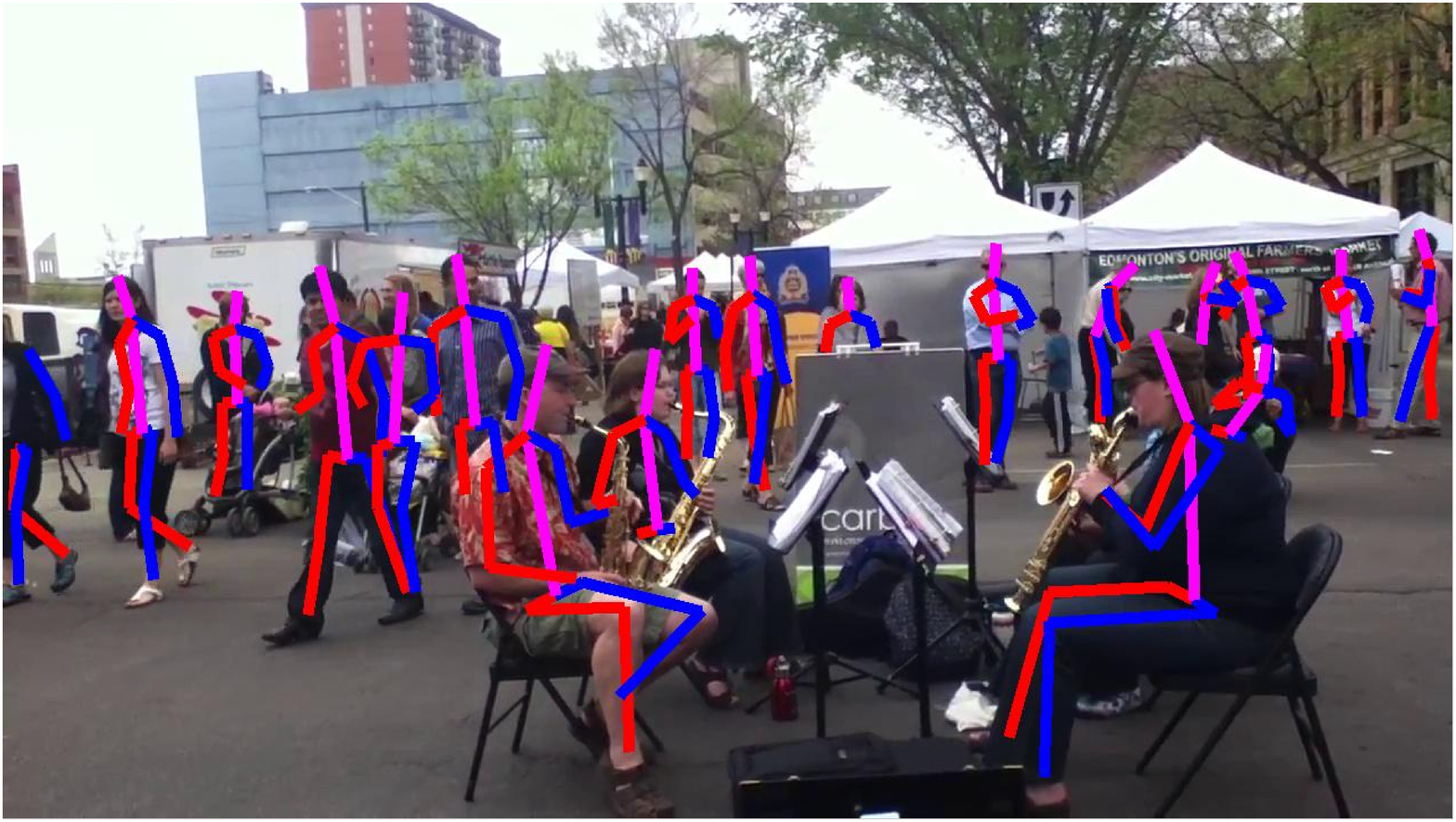}&
\includegraphics[width=0.25\linewidth]{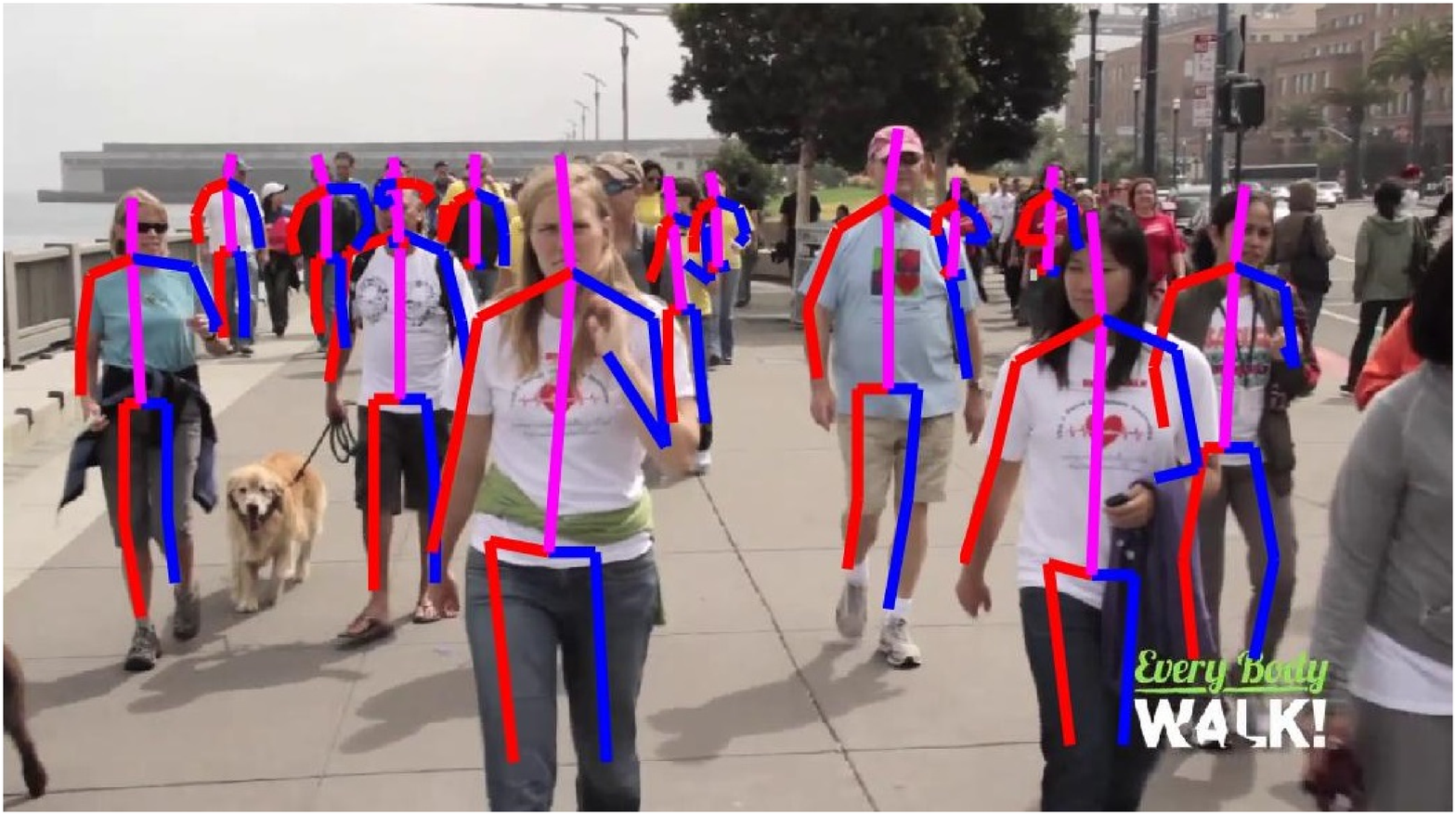}&
\includegraphics[width=0.25\linewidth]{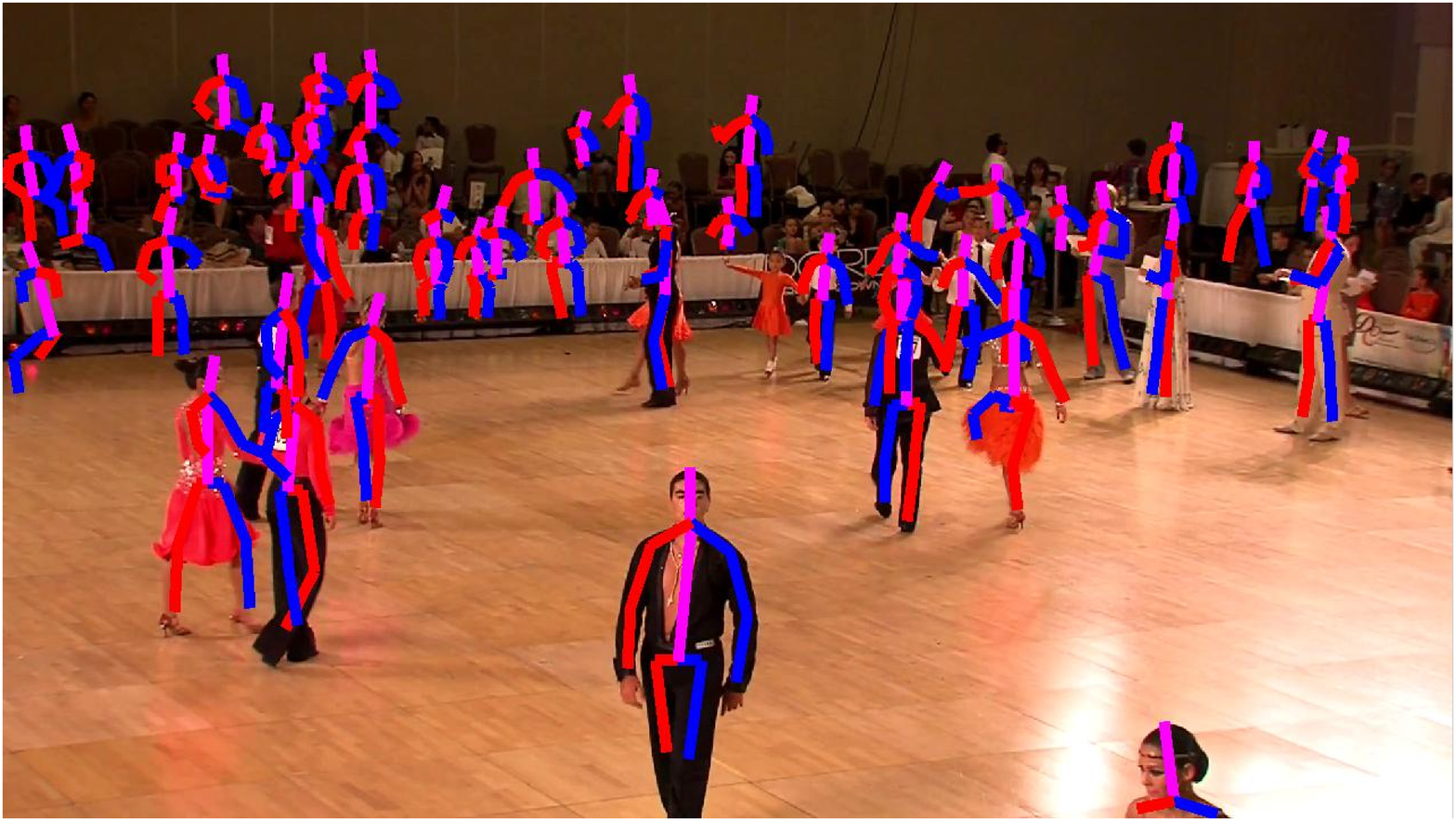}&
\includegraphics[width=0.25\linewidth]{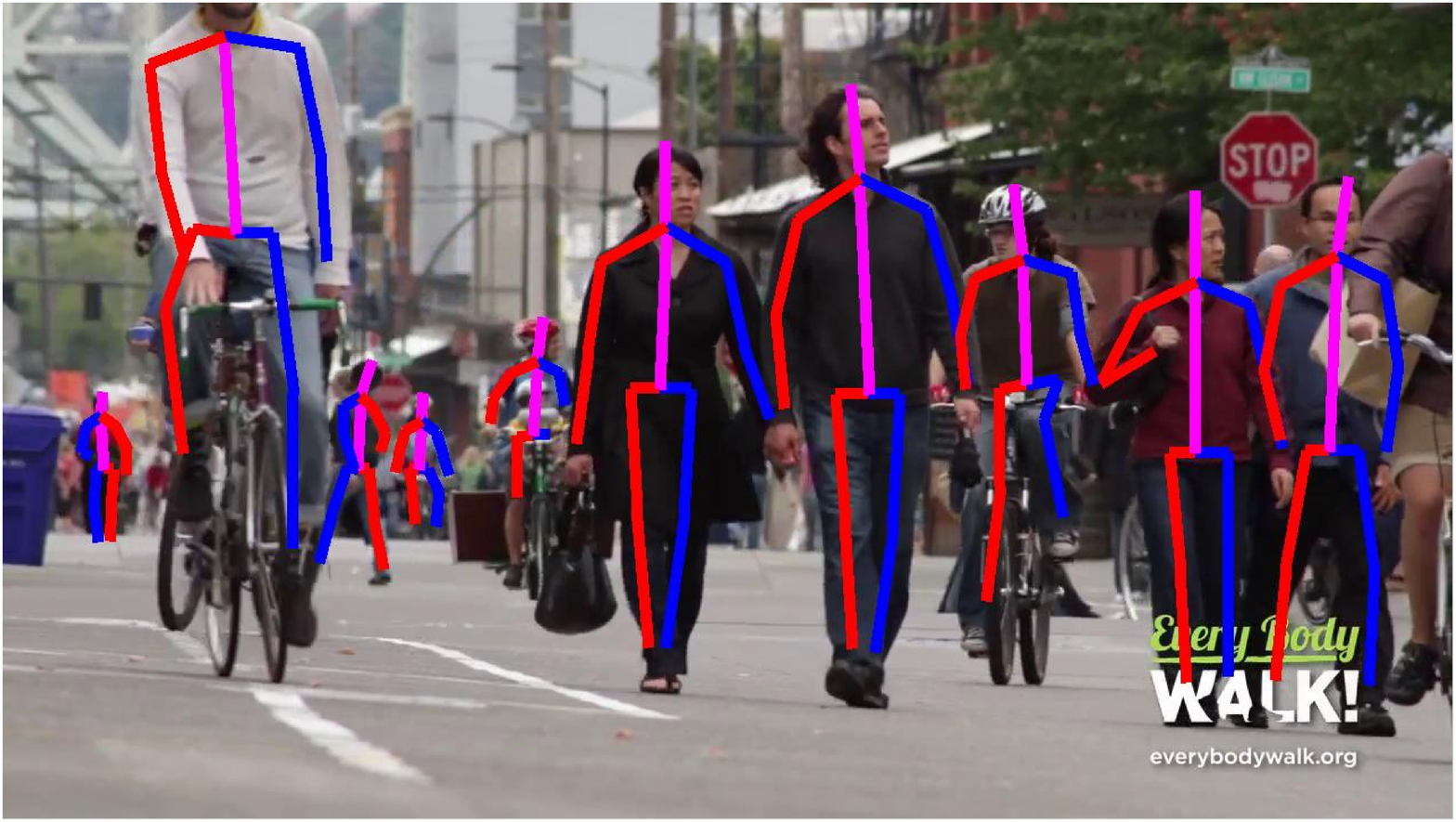}\\
\includegraphics[width=0.25\linewidth]{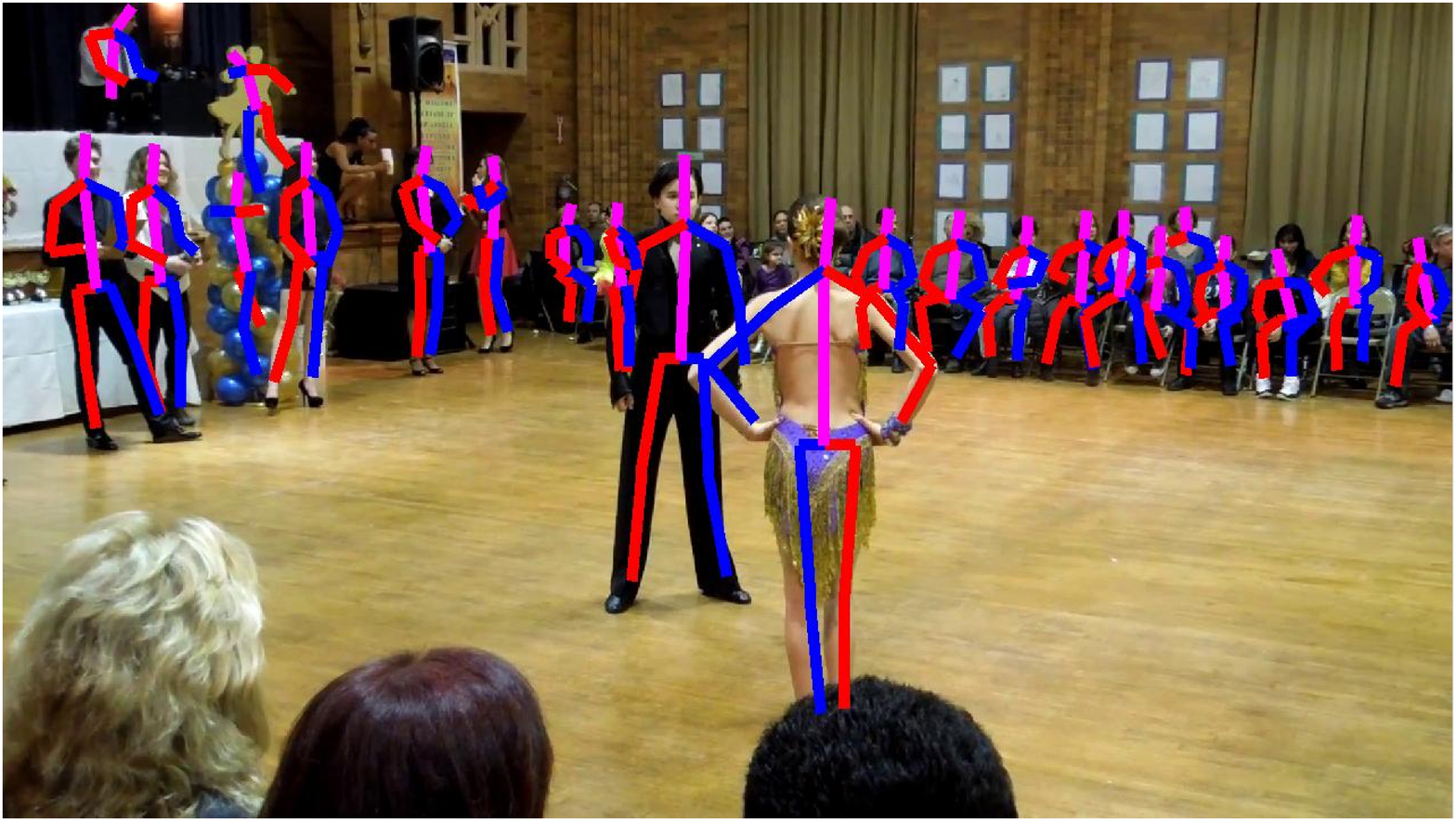}&
\includegraphics[width=0.25\linewidth]{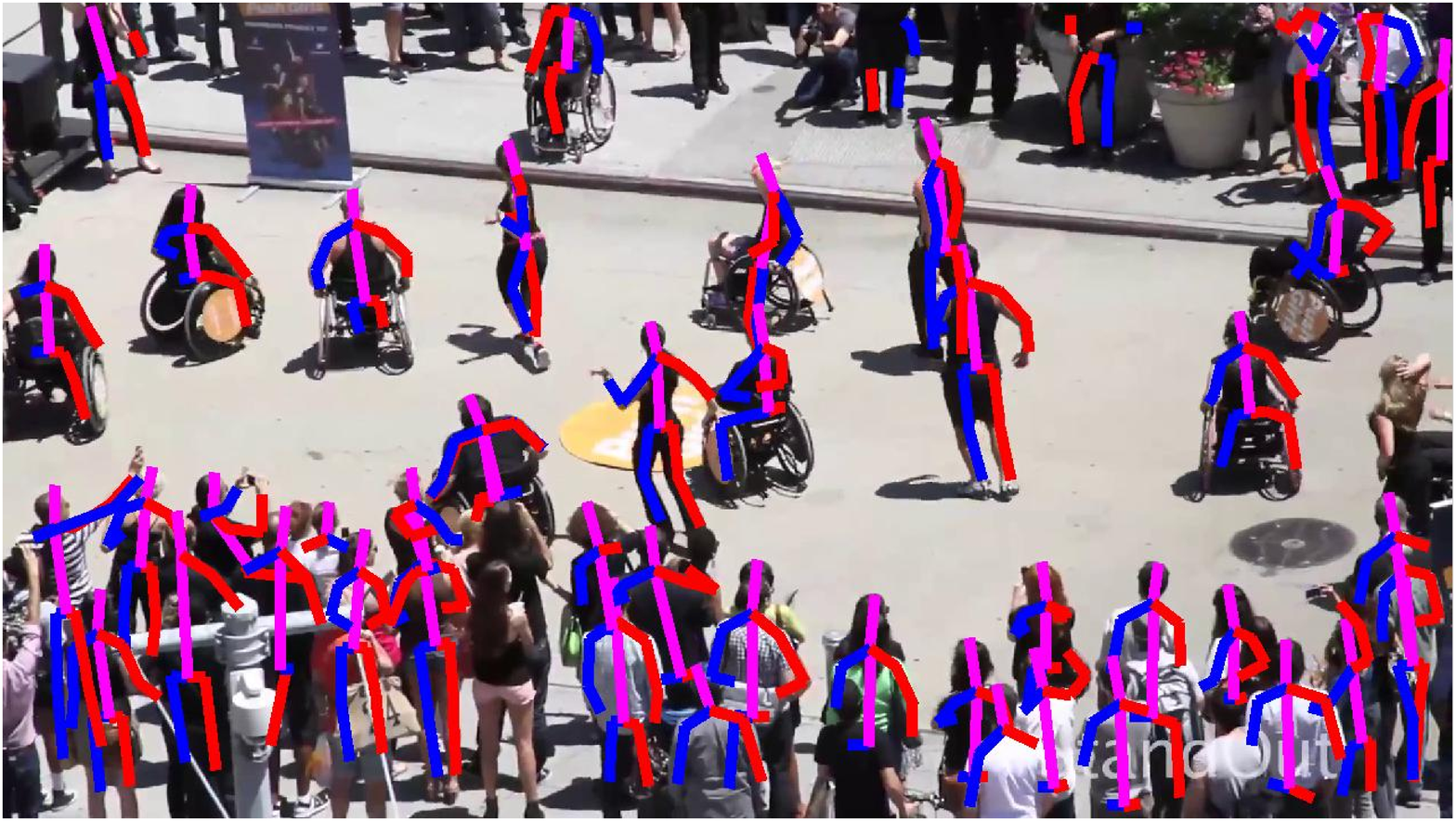}&
\includegraphics[width=0.25\linewidth]{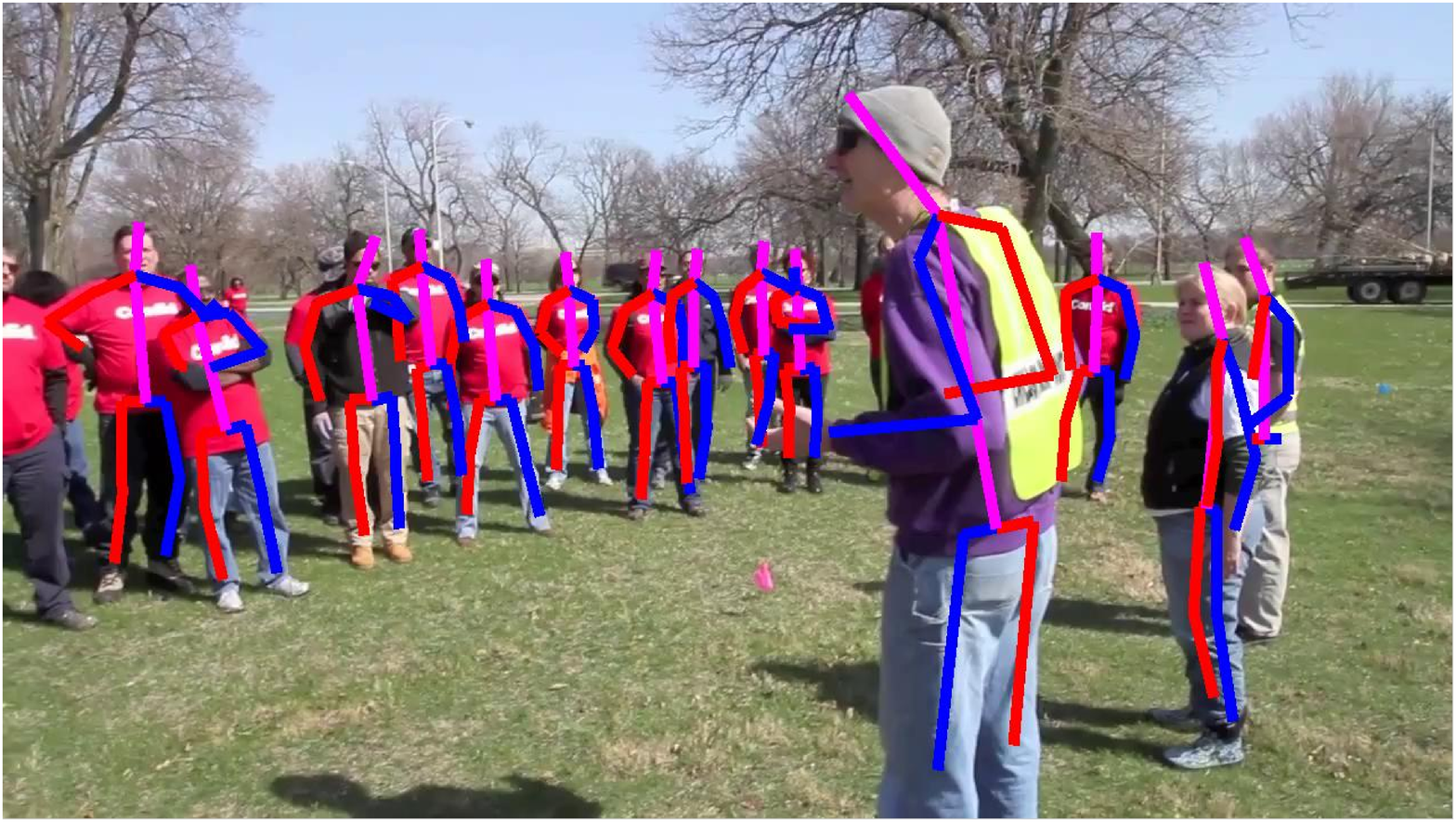}&
\includegraphics[width=0.25\linewidth]{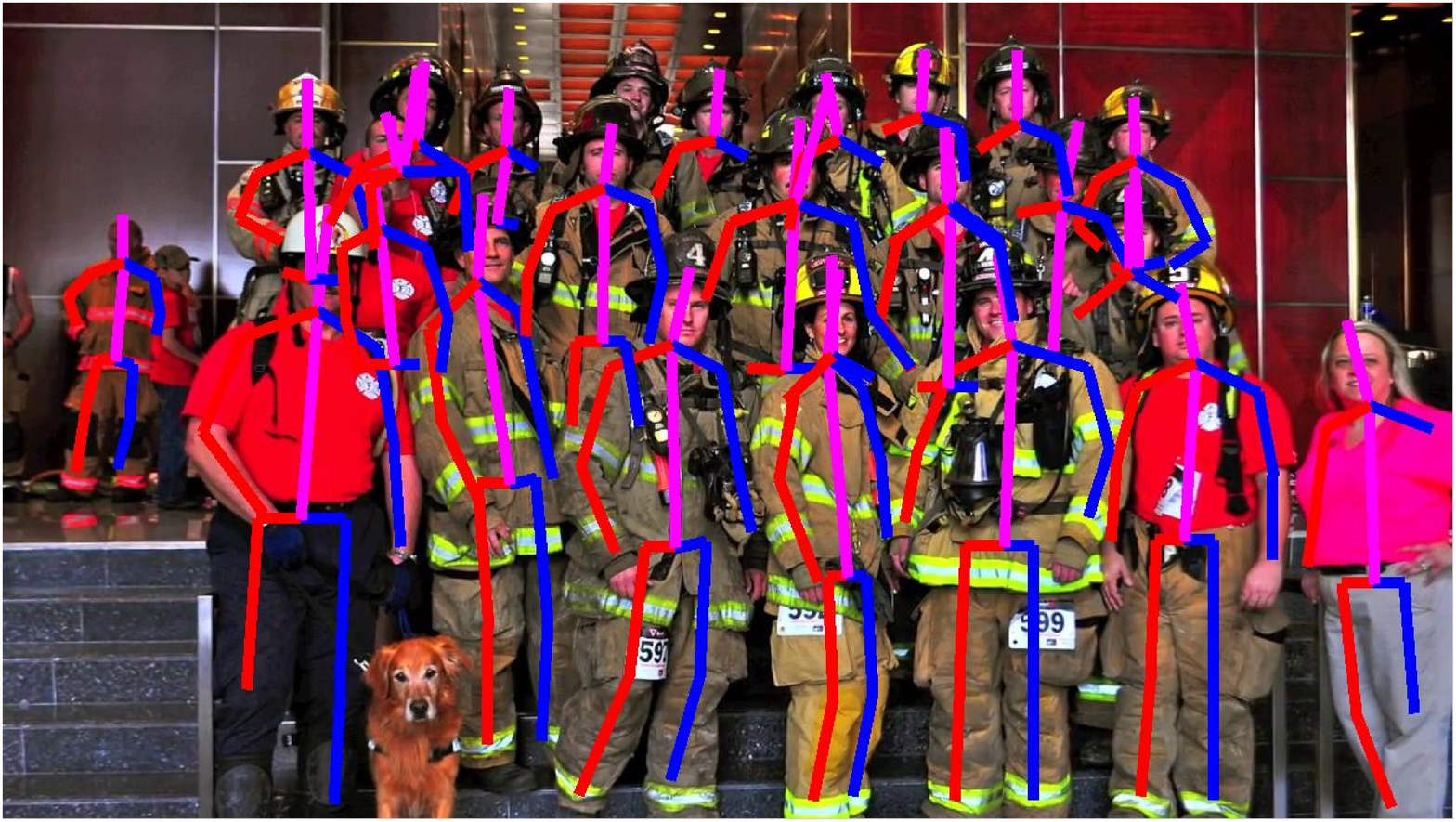}

\end{tabular}
\caption{Some results of our model's predictions.}\vspace{-0.10in}
\label{fig:res}
\end{figure*}

\begin{figure*}[hbt]
\centering
\begin{tabular}{@{\hspace{0mm}}c@{\hspace{1mm}}c@{\hspace{1mm}}c@{\hspace{1mm}}c}
\includegraphics[width=0.25\linewidth]{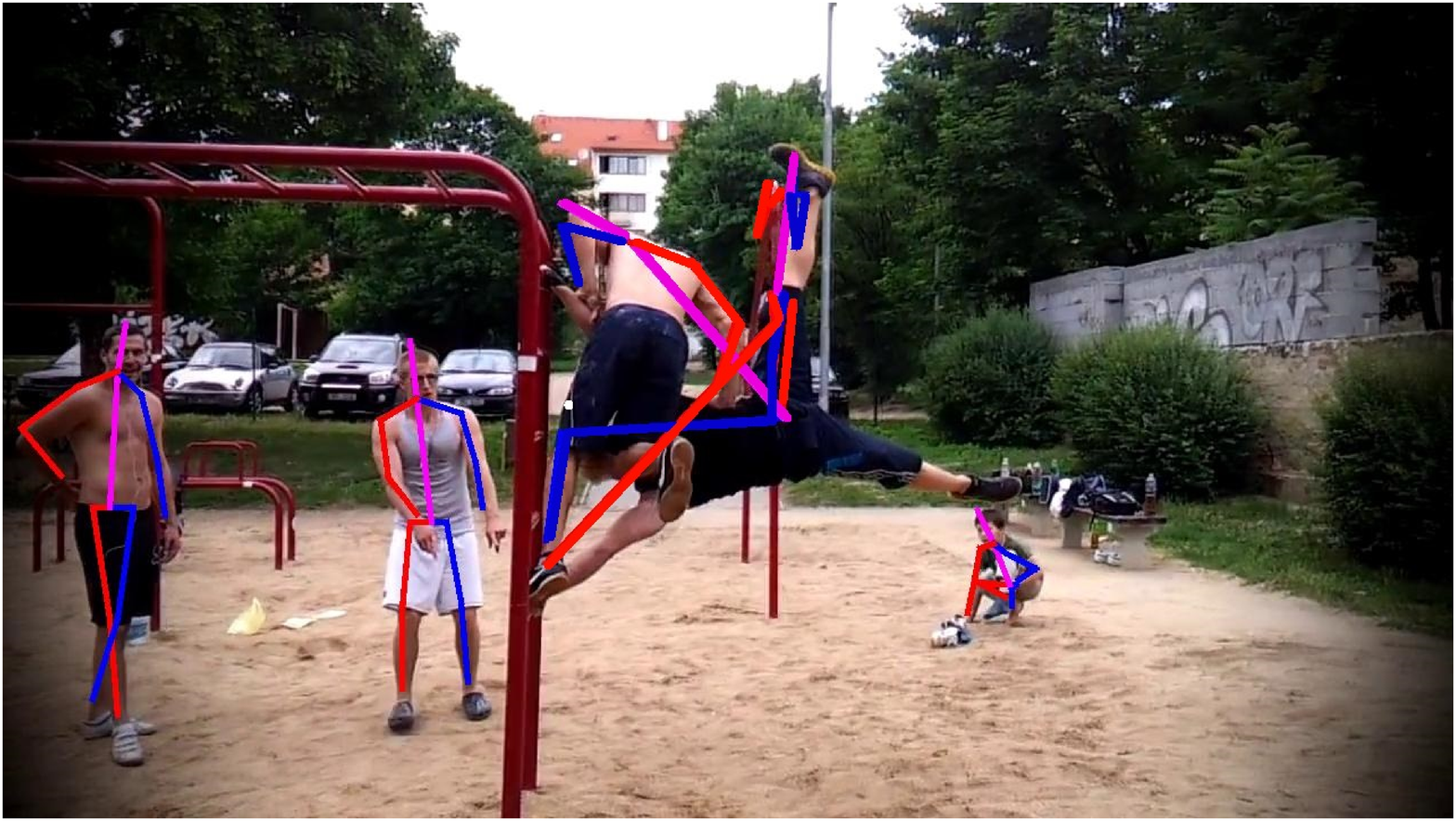}&
\includegraphics[width=0.25\linewidth]{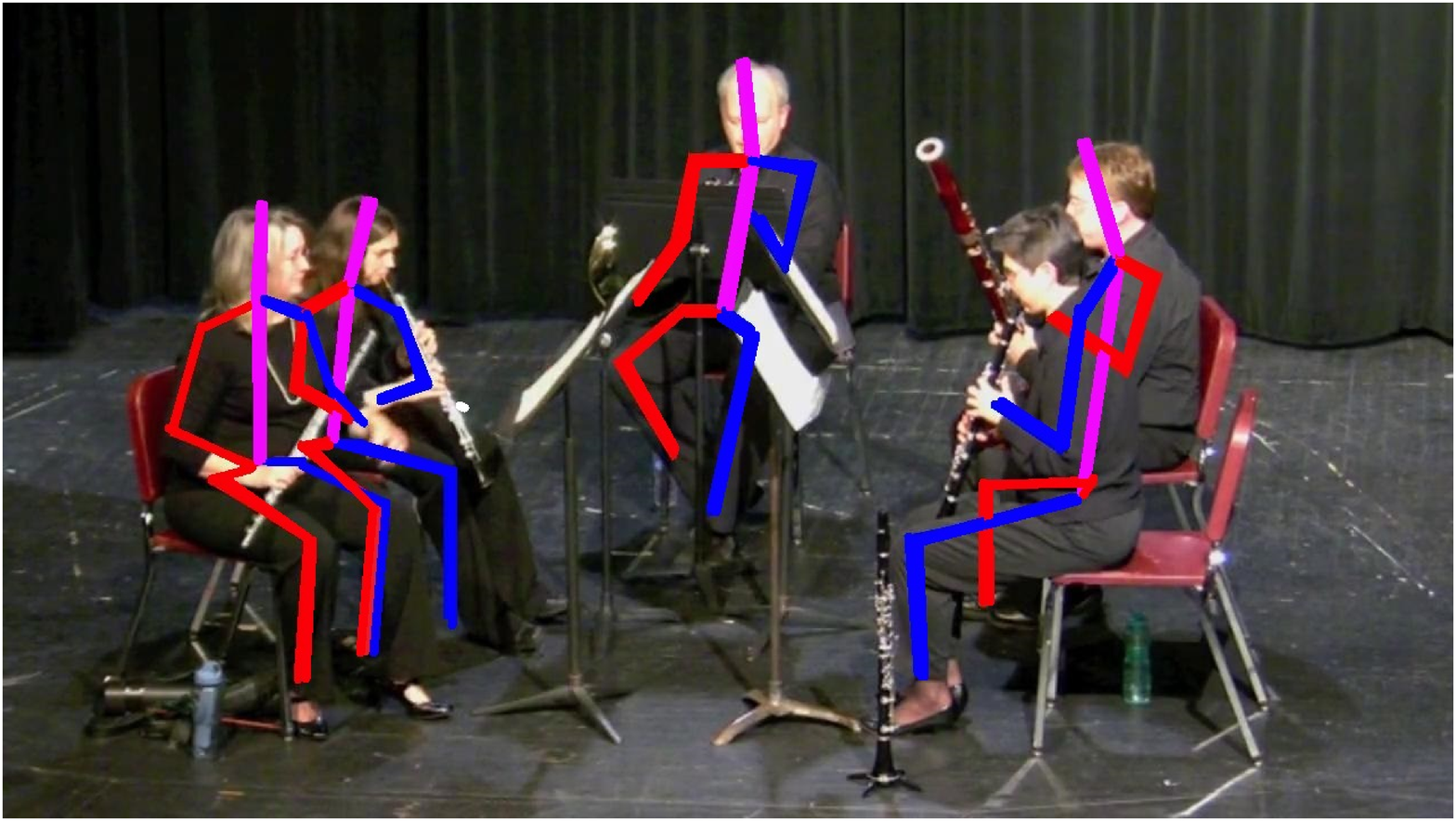}&
\includegraphics[width=0.25\linewidth]{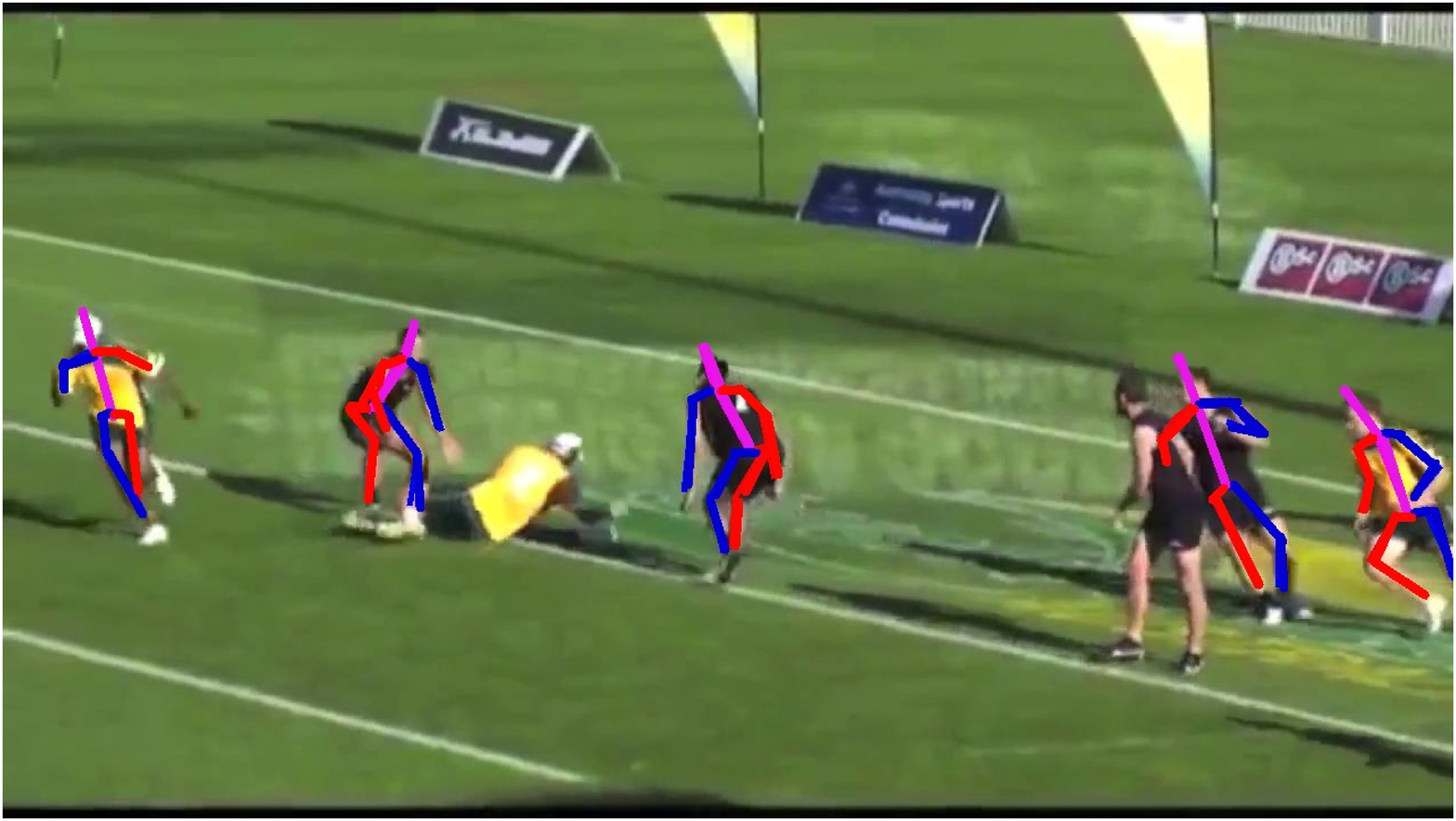}&
\includegraphics[width=0.25\linewidth]{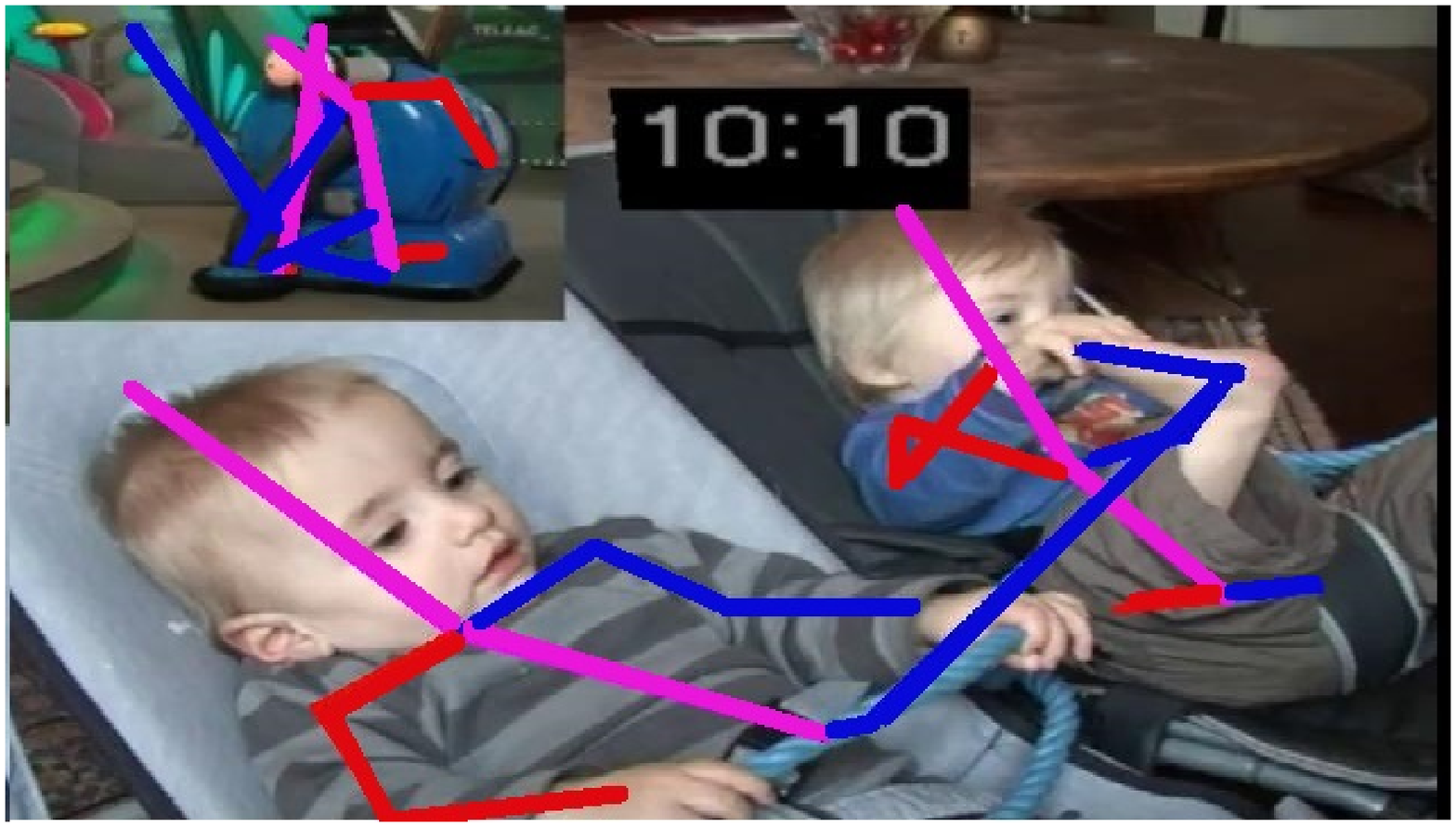}

\end{tabular}
\caption{Example failure cases of our model.}\vspace{0.0in}
\label{fig:resfail}
\end{figure*}
\paragraph{Results on MPII dataset.} We evaluated our method on full MPII multi-person test set. Quantitative results on the full testing set are given in Table \ref{tab:compare}.
Notably, we achieve an average accuracy of $72$ mAP on identifying difficult joints such as wrists, elbows, ankles, and knees, which is $3.3$ mAP higher than the previous  state-of-the-art result. We reach a final accuracy of $70.4$ mAP for the wrist and an accuracy of $73$ mAP for the knee. By using a stronger human detector and pose estimator, we can further achieve $82.1$ mAP, which is $4.6$ mAP higher than the previous best result. We present some of our results in Figure~\ref{fig:res}. These results show that our method can accurately predict pose in multi-person images. More results are presented in supplementary materials.


\vspace{2mm}
\noindent{\bf Results on MSCOCO Keypoints dataset.} We fine-tuned the SPPE on the MSCOCO Keypoints training + validating sets and leave 5,000 images for validation. Quantitative results on the test-dev set are given in Table \ref{tab:compareCOCO}. Our method achieves the state-of-the-art performance. Note that without specific design for the pose estimation network, our frame work can perform on par with Megvii\cite{chen2017cascaded}, which propose a new pose estimation network. It demonstrates the effectiveness of our proposed framework. And we believe that using the pose network from~\cite{chen2017cascaded} can further boost our performance.
{\small
\begin{table}[tbh]
\begin{center}
\resizebox{1\linewidth}{!}{
\begin{tabular}{c|c |c c c c}
\hline
Team &\textbf{AP} & $AP^{50}$ & $AP^{75}$ & $AP^{M}$  & $AP^{L}$\\
 \hline
CMU-Pose\cite{cao2017realtime}&  61.8  &  {84.9} &   67.5 &  57.1&   {68.2}  \\
G-RMI\cite{papandreou2017towards} &68.5  & 87.1&  75.5 &  65.8  &  73.3 \\
Mask R-CNN\cite{he2017mask} &63.1 &87.3 &68.7 &57.8 &71.4 \\
Megvii\cite{chen2017cascaded} &72.1 &91.4 &80.0 &68.7 &77.2 \\
\hline
ours & 61.8  & 83.7  & {69.8}  & {58.6}  & 67.6\\
ours++ & \textbf{72.3}  & 89.2  & {79.1}  & 68.0  & \textbf{78.6}\\
\hline
\end{tabular}}
\vspace{0.10in}
\caption{Results on the MSCOCO Keypoint Challenge (AP) dataset \cite{MSCOCO}. The MSCOCO website provides a technical overview only. Our result is obtained without ensembling. ``++'' denotes using faster-rcnn with softnms~\cite{bodla2017soft} as human detector, PyraNet~\cite{yang2017learning} with input size 320x256 as pose estimator. We only compare to single model results.} \label{tab:compareCOCO}
\end{center}
\end{table}
}

\subsection{Ablation studies}
We evaluate the effectiveness of the three proposed components, i.e., symmetric STN, pose-guided proposals generator and parametric pose NMS. The ablative studies have been conducted by removing the proposed components from the pipeline or replacing the proposed components with conventional solvers. The straightforward two-step method without the three components and the upper-bound of our framework are tested for comparison. We conducted these experiments on the MPII validation set. In addition, we replace our human detection module to prove the generality of our framework.
\begin{table*}[tbh]
\begin{center}
\begin{tabular}{ll|c c c c c c c c c}
\hline
 &Methods &Head & Shoulder & Elbow & Wrist & Hip & Knee  & Ankle & Total\\
 \hline
& \textbf{RMPE, full}& \textbf{90.7} & \textbf{89.7}  & \textbf{84.1}  & \textbf{75.4}  & \textbf{80.4}  & \textbf{75.5} & \textbf{67.3} & \textbf{80.8}\\ \hline
\multirow{2}{*}{a)} & w/o SSTN+parallel SPPE& 89.0  & 86.9  & 82.8  & 73.5  & 77.1  & 73.3 & 65.0 & 78.2 \\
& w/o parallel SPPE only& 89.9  & 88.0  & 83.4  & 74.7  & 77.8  & 74.0 & 65.8 & 79.1 \\
\multirow{2}{*}{b)} & w/o PGPG&  82.8  & 81.0  & 77.5  & 68.2  & 74.6  & 66.8 & 60.1 & 73.0 \\
& random jittering* &  89.3  & 87.8  & 82.3  & 70.4  & 78.4  & 73.3 & 63.8 & 77.9 \\
\multirow{3}{*}{c)}& w/o PoseNMS& 85.1  & 83.6  & 79.2  & 69.8  & 76.4  & 72.2 & 63.6 & 75.7 \\
& PoseNMS \cite{chen2015parsing}& 88.9  & 87.8  & 83.0  & 73.8  & 78.7  & 74.6 & 66.3 & 79.1\\
& PoseNMS \cite{burgos2013merging}& 90.0  & 88.6  & 83.7  & 74.6  & 79.7  & 75.1 & 67.0 & 79.9\\
d)& straight forward two-steps & 81.9  & 80.4  & 74.1  & 68.5  & 69.0  & 66.1 & 62.2 & 71.7 \\
e)& oracle human detection & 94.3  & 93.4  & 87.7  & 80.2  & 84.3  & 78.9 & 70.6 & 84.2 \\
\end{tabular} \vspace{0.10in}
\caption{Results of the ablation experiments on our validation set. ``w/o X'' means without X module in our pipeline. ``random jittering*'' means generating training proposals by jittering locations and aspect ratios of the detected human bounding boxes. ``PoseNMS [x]'' reports the result when using the pose NMS algorithm developed in paper [x].} \label{tab:Ablation}
\end{center}
\vspace{-0.10in}
\end{table*}

\vspace{1mm}
\noindent{\bf Symmetric STN and Parallel SPPE} To validate the importance of symmetric STN and parallel SPPE, two experiments were conducted. In the first experiment, we removed the SSTN, including the parallel SPPE, from our pipeline. In the second experiment, we only removed the parallel SPPE and kept the symmetric STN structure. Both of these results are shown in Table \ref{tab:Ablation}(a). We can observe performance degradation when removing parallel SPPE, which implies that parallel SPPE with single person image labels strongly encourages the STN to extract single person regions to minimize the total losses.

\vspace{1mm}
\noindent{\bf Pose-guided Proposals Generator} In Table \ref{tab:Ablation}(b), we demonstrate that our pose-guided proposals generator also plays an important role in our system. In this experiment, we first remove the data augmentation from our training phase. The final mAP drops to $73.0\%$. Then we compare our data augmentation technique with a simple baseline. The baseline is formed by jittering the locations and aspect ratios of the bounding boxes produced by person detector to generate a large number of additional proposals. We choose those that have IoU$>$0.5 with ground truth boxes. From our result in Table \ref{tab:Ablation}(b), we can see that our technique is better than the baseline method. Generating training proposals according to the distribution can be regarded as a kind of data re-sampling, which can help the model to better fit human proposals.

\vspace{1mm}
\noindent{\bf Parametric Pose NMS} Since pose NMS is an independent module, we can directly remove it from our final model. The experimental results are shown in Table \ref{tab:Ablation}(c). As we can see, the mAP drops significantly if the parametric pose NMS is removed. This is because the increase in the number of redundant poses will ultimately decrease our precision. We note that the previous pose NMS can also eliminate redundant detection to some extent. The state-of-the-art pose NMS algorithms \cite{burgos2013merging,chen2015parsing} are used to replace our parametric pose NMS, with the results given in Table \ref{tab:Ablation}(c). These schemes perform less effectively than ours, since the parameter learning is missing. In terms of efficiency, on our validation set which contains 1300 images, the publicly available implementation of \cite{burgos2013merging}\footnotemark[3]takes 62.2 seconds to perform pose NMS while using our algorithm takes only 1.8 seconds.
\footnotetext[3]{\href{http://www.vision.caltech.edu/~dhall/projects/MergingPoseEstimates/}{http://www.vision.caltech.edu/~dhall/projects/MergingPoseEstimates/}}

\vspace{1mm}
\noindent{\bf Upper Bound of Our Framework} The upper bound of our framework is tested, where we use the ground truth bounding boxes as human proposals. As shown in Table \ref{tab:Ablation}(e), this setting could yield $84.2\%$ mAP. It verifies that our system is already close to the upper-bound of two-step framework.

\subsection{Failure cases}
We present some failure cases in Figure~\ref{fig:resfail}. It can be seen that the SPPE can not handle poses which are rarely occurred (e.g. the person performing the 'Human Flag' in the first image). When two persons are highly overlapped, our system get confused and can not separate them apart (e.g. the two persons in the left of the second image). The misses of person detector will also cause the missing detection of human poses (e.g. the person who has laid down in the third image). Finally, erroneous pose may still be detected when an object looks very similar to human  which can fool both human detector and SPPE (e.g. the background object in the forth image).

\section{Conclusion}
In this paper, a novel regional multi-person pose estimation (RMPE) framework is proposed, which significantly outperforms the state-of-the-art methods for multi-person human pose estimation in terms of accuracy and efficiency. It validates the potential of two-step frameworks, i.e., human detector + SPPE, when SPPE is adapted to a human detector. Our RMPE framework consists of three novel components: symmetric STN with parallel SPPE, parametric pose NMS, and pose-guided proposals generator (PGPG). In particular, PGPG is used to greatly argument the training data by learning the conditional distribution of bounding box proposals for a given human pose. The SPPE becomes adept at handling human localization errors due to the utilization of symmetric STN and parallel SPPE. Finally, the parametric pose NMS can be used to reduce redundant detections.
In our future work, it would be interesting to explore the possibility of training our framework together with the human detector in an end-to-end manner.

%
%
%

{\small
\bibliographystyle{ieee}
\bibliography{egbib}
}

\end{document}